 \documentclass[final,3p,times]{elsarticle}

\usepackage{url}

\usepackage{multirow}

\usepackage{adjustbox}
\usepackage[utf8]{inputenc}

\usepackage{graphicx}
\usepackage{multirow}%
\usepackage{amsmath,amssymb,amsfonts}%
\usepackage{amsthm}%
\usepackage{mathrsfs}%
\usepackage[title]{appendix}%

\usepackage[dvipsnames]{xcolor}
\usepackage{xcolor}%
\usepackage{textcomp}%
\usepackage{manyfoot}%
\usepackage{booktabs}%
\usepackage{algorithm}%
\usepackage{algorithmicx}%
\usepackage{algpseudocode}%
\usepackage{listings}%
\usepackage{natbib}%
\usepackage{tabularx}
\usepackage{tabulary}
\usepackage{threeparttable} 
\usepackage{comment}

\usepackage{hyperref}

\usepackage{times}
\usepackage{epsfig}
\usepackage{multicol}
\usepackage{makecell}
\usepackage{enumitem}
\usepackage{array}
\usepackage{siunitx}
\usepackage{mathtools}
\usepackage{nomencl}
\usepackage{soul}
\usepackage[usestackEOL]{stackengine}
\usepackage{soul}

\usepackage{listings}
\definecolor{codegreen}{rgb}{0,0.6,0}
\definecolor{codegray}{rgb}{0.5,0.5,0.5}
\definecolor{codepurple}{rgb}{0.58,0,0.82}
\definecolor{backcolour}{rgb}{0.95,0.95,0.92}

\lstdefinestyle{mystyle}{
    backgroundcolor=\color{backcolour},   
    commentstyle=\color{codegreen},
    keywordstyle=\color{magenta},
    numberstyle=\tiny\color{codegray},
    stringstyle=\color{codepurple},
    basicstyle=\ttfamily\footnotesize,
    breakatwhitespace=false,         
    breaklines=true,                 
    captionpos=b,                    
    keepspaces=true,                 
    numbers=left,                    
    numbersep=5pt,                  
    showspaces=false,                
    showstringspaces=false,
    showtabs=false,                  
    tabsize=2
}
\usepackage{kotex}

\usepackage[capitalize]{cleveref}
\crefname{section}{Sec.}{Secs.}
\Crefname{section}{Section}{Sections}
\Crefname{table}{Table}{Tables}
\crefname{table}{Tab.}{Tabs.}

\newcommand{\norm}[1]{\| #1 \|}
\newcommand{\bignorm}[1]{\Bigl \| #1 \Bigr \| }

\begin{document}

\begin{frontmatter}

\title{Selective Correlation Based Knowledge Distillation \\ for Ground Reaction Force Estimation}

\author[label1]{Eun~Som~Jeon}
\author[label2]{Jisoo Lee}
\author[label1]{Huisu Lim}
\author[label3]{Omik M. Save}
\author[label3]{Hyunglae Lee}
\author[label4]{Pavan~Turaga}

\affiliation[label1]{organization={Department of Computer Science and Engineering, Seoul National University of Science and Technology},
             city={Seoul},
             postcode={01811},
             country={Republic of Korea}}
\affiliation[label2]{organization={Geometric Media Lab, School of Computing and Augmented Intelligence, Arizona State University},
             city={Tempe},
             postcode={85281},
             state={AZ},
             country={USA}}


\affiliation[label3]{organization={School for Engineering of Matter, Transport and Energy, Arizona State University},
             city={Tempe},
             postcode={82587},
             state={AZ},
             country={USA}}

\affiliation[label4]{organization={Geometric Media Lab, The GAME School and School of Electrical, Computer and Energy Engineering: Home
, Arizona State University},
             city={Tempe},
             postcode={85281},
             state={AZ},
             country={USA}}
             

\begin{abstract}
Wearable sensor-based human gait analysis holds great promise in healthcare, rehabilitation, clinical diagnosis and monitoring, and sports activities. Specifically, ground reaction force (GRF) provides essential insights into the body's interaction with the ground during movement and is typically measured using instrumented treadmills equipped with force plates. However, such equipment is expensive and restricted to laboratory environments. To enable a more portable solution, wearable insole sensors have been used to measure GRF. These sensors, however, are prone to noise and external interference, which reduces measurement accuracy. Deep learning methodologies could be adopted to address these issues, but they often require significant computing resources to achieve high accuracy, limiting their applicability for real-time analysis on portable devices. To overcome these limitations, we propose Selective Correlation Based Knowledge Distillation (SCKD) for estimating GRF from data collected by insole sensors. Our proposed method utilizes selected features considering temporal characteristics in the process of extracting correlation maps for knowledge transfer, enhancing interpretability and mitigating issues in high dimensional data processing. We demonstrate the effectiveness of the compact models generated by our distillation framework through comparison with existing methods. Various configurations of teacher-student architectures and training approaches are examined based on multiple evaluation criteria, utilizing data collected at different walking speeds and with different window sizes. Experimental results confirm that our approach outperforms existing methods in estimating GRF from wearable insole sensor data. Therefore, our approach offers a reliable and resource-efficient solution for human gait analysis. 
\end{abstract}

\begin{keyword}
Ground reaction force, insole sensor, knowledge distillation, wearable sensor data, sensor data estimation. 


\end{keyword}

\end{frontmatter}



\section{Introduction}\label{sec1}
The integration of human-centric Internet of Things (IoT) utilizing wearable devices has increasingly attracted interest due to various practical applications including healthcare oriented daily life monitoring, physical activity tracking, and smart home implementations \cite{chen2019bring, cai2022mhealth, dang2020sensor, yang2018smart}. Specifically, gait and movement analyses are essential in diagnosing movement disorders such as hemiplegia \cite{brunelli2020early} and Parkinson’s disease \cite{li2021rehabilitation}, which are involved in maintaining balance, resulting in unstable and irregular gait patterns. Furthermore, wearable derived gait signals facilitate injury prevention and rehabilitation by tracking patient intent and recovery, offering significant data to control exoskeleton robotics in therapeutic scenarios \cite{rajasekaran2018volition, su2023review}.


For movement assessment and activity monitoring, gait analysis using ground reaction force (GRF) sensors has been widely employed \cite{xiang2024rethinking, buurke2023comparison}, particularly in combination with deep learning approaches \cite{buurke2023comparison, an2023artificial}. GRF represents the reactive force exerted on the foot during ground contact \cite{Key2010}. Since GRF is significantly influenced by human movement, this can be utilized to accurately predict lower limb motions during dynamic tasks \cite{Sakamoto2023}. 
While instrumented treadmills effectively measure GRF during walking or running \cite{kluitenberg2012comparison}, their practicality in real-time and everyday monitoring systems is limited due to expensive instrumentation and limited accessibility for most individuals \cite{howell2013kinetic}. Moreover, utilizing multiple sensors simultaneously during test time substantially increases system complexity and implementation cost \cite{shi2023robust}.

Recent studies have suggested utilizing in-shoe systems to mitigate these limitations and estimate GRF effectively \cite{burns2019validation, lee2019functional}. For instance, the wireless insole-based plantar load measurement system, Loadsol (Novel GmbH, Germany), demonstrated measurement performance comparable to treadmill-based force plates \cite{burns2019validation}. The Tactilus High-Performance V-Series (SensorProd Inc., USA) insole offer dense sensor arrays over 100 sensing points, however, there are issues related to shoe size and signal drift under constant force over a short period, resulting in considerable errors \cite{chen2024center, Gehlhar2022}. Low-quality sensing units further limit analytical performance, notably during stance phase of initiation and termination \cite{burns2019validation, dyer2011instrumented}. Additional errors arise from system instability, personal differences, and extrinsic gait variability \cite{masani2002variability}.

To enhance signal quality and accuracy, deep learning models (e.g. autoencoder) have been utilized \cite{Hinton2006Dim}. Typically, deeper networks improve performance but also increase trainable parameters, increasing computational complexity and storage resources \cite{huang2017densely, fawaz2019deep, khan2020survey}. With the increasing demand for real-time data processing in wearable sensor analyses, developing lightweight models has become imperative, which maintain high performance while requiring fewer resources \cite{saad2024employing, ni2024survey}. On the other hand, traditional deep learning methods in estimation tasks generally operate in a unimodal manner \cite{Li2022CNN, Lipton2015RNN}, where input and output share similar statistical properties and data types. When input (video data) and output (time-series data) have different data types in a multimodal manner, the model performance typically is degraded, which is trained in a unimodal manner.

To overcome these limitations, we propose a robust framework in knowledge distillation (KD) to effectively estimate GRF from insole sensor data, which generates a lightweight model. Conventional KD is to create a small model (student) by knowledge of a large model (teacher) \cite{hinton2015distilling}, whose effectiveness has been demonstrated in wearable sensor data analysis \cite{jeon2022kd, gou2021knowledge}. An important consideration for making better students in distillation involves the information provided by teacher models. To use superior knowledge in distillation, previous studies utilize correlations in knowledge transfer, representing similarities among samples within a mini-batch \cite{tung2019similarity}. However, high-dimensional features generally do not fully capture the relationships between individual elements, and the computational complexity of analyzing these relationships is also very high. Thus, sparse correlation or dimension reduction is generally leveraged in machine learning to reduce noise and improve the interpretation of features in processing \cite{johnstone2009statistical}.

\begin{figure*}[t]
\includegraphics[width=0.99\linewidth]{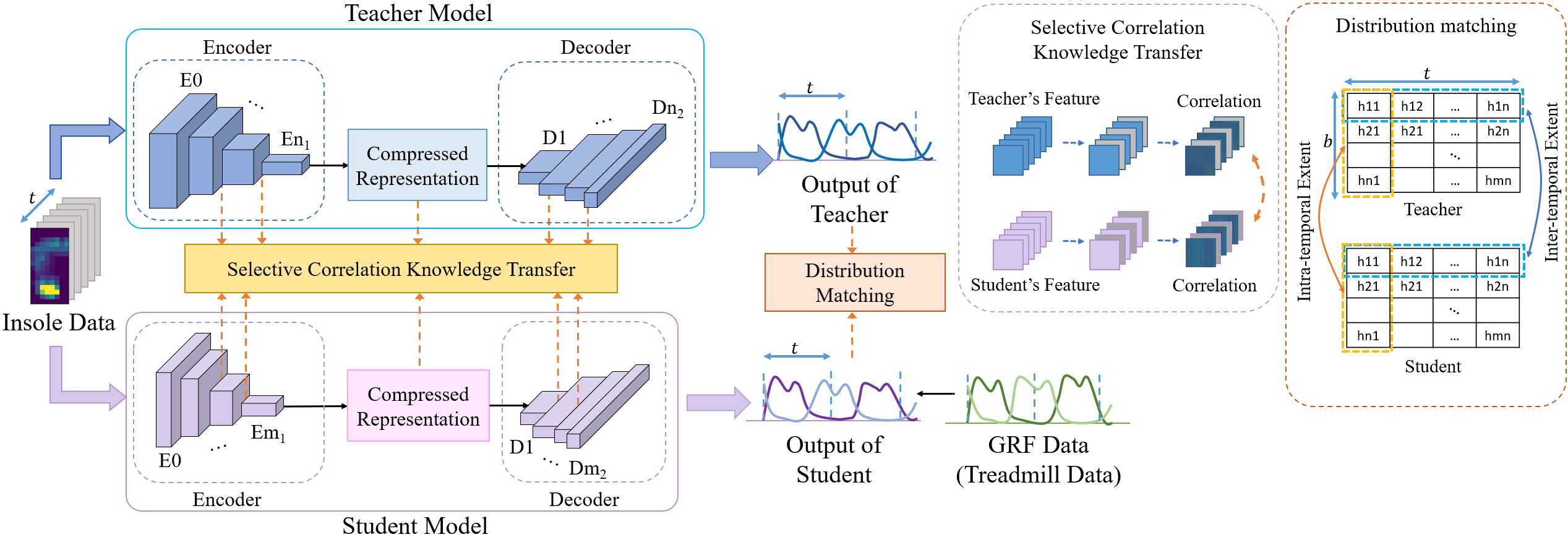} 
\centering
\caption{Overview of Selective Correlation Based Knowledge Distillation, SCKD.}
\label{figure:overview}
\end{figure*}

In this context, we devise a framework that we term Selective Correlation Based Knowledge Distillation, SCKD, illustrated in Fig. \ref{figure:overview}. For features in intermediate layers, small number of channels are chosen and their features are used in computation of correlation for knowledge transfer, rather than using all possible features. Since our target is time-series estimation, preserving temporal characteristics is crucial in model learning. To consider temporal information, the selected feature maintains the size of temporal extent. This reduces complexity of features and increases interpretability in correlation extraction. Furthermore, this provides more room to encourage the student model to train itself in a way that suits its own architecture.
Additionally, to mimic performance of the teacher thoroughly, we applied distribution matching algorithm in comparison between the model outputs, which provides strong supervision in estimation task. This utilizes probabilistic distribution of inter- and intra-temporal extents within a mini-batch, which encode relationships of the same time intervals and temporal trends over successive time intervals. Finally, this process yields a compact student model that surpasses equally-sized models distilled by other methods.

For our experiments to meet our specified requirements, we designed a customized system architecture to gather comprehensive GRF data from instrumented treadmills and insole sensors, enabling flexible experimentation across various configurations. Utilizing this dataset, we demonstrate performance of our framework considering different window sizes and data formats.

The primary contributions of our work include:
\begin{itemize}
\item We propose a novel knowledge distillation framework, which is to generate a robust lightweight model through knowledge transfer from a larger model for GRF estimation.
\item We introduce an effective knowledge transfer approach leveraging selective feature correlation for improved GRF estimation, leveraging multimodal data as input and output. 
Also, we introduce inter- and intra-temporal feature matching method, which effectively matches feature distributions between teacher and student models.
\item We develop a flexible system architecture tailored for precise insole sensor-based GRF estimation. This enables us to investigate the distilled models with various evaluations across varying data scales and window size.
\item We rigorously demonstrate the robustness of our approach with diverse empirical assessments, comparing with various KD methods, validating on multiple metrics, different teacher-student configurations, different learning methods, and reliability.
\end{itemize}

The remainder of the paper is structured as follows: Section \ref{sec:background} outlines background knowledge, including systems and deep learning methods. Section \ref{sec:proposed_method} describes our proposed system architecture, dataset preparation, and the proposed KD framework. Sections \ref{sec:experiments} and \ref{sec:ablations} present experimental results and their analyses, followed by discussion and conclusion in Sections \ref{sec:discussion} and \ref{sec:conclusion}.


\section{Background} \label{sec:background}
\subsection{Deep Networks for Estimation}

Many works have explored multi-channel sensor data fusion with deep learning techniques \cite{Zhang2023inception, Tian2019fusion}, usually for image or action classification tasks. In biomechanics related works, there have been efforts at estimating the Center of Pressure (CoP) from individual video frames, but the methods did not fully leverage temporal information \cite{Chen2024cop}. 
Another work compared temporal modeling abilities of 3D CNNs and C-LSTMs without consideration of real-time processing efficiency \cite{Manttari_2020_ACCV}.
There have also been many efforts at computational cost reduction in sensor data processing using knowledge distillation approaches \cite{Mardanpour2023activity, Jeong2022depth}. 

For learning representations and addressing estimation challenges, 
Autoencoders (AE), Variational Autoencoders (VAE) \cite{kingma2013auto,fabius2015variational}, and Wasserstein Autoencoders (WAE) \cite{tolstikhin2018wasserstein} have been extensively employed.
AEs are one of the most popular methods in representation learning by transforming high-dimensional inputs into lower-dimensional latent spaces (denoted as $\mathcal{Z}$). An AE typically comprises an encoder ($E(\cdot)$), which projects input data to latent representations $\mathcal{Z}$, and a decoder ($D(\cdot)$) reconstructing outputs from these latent representations. The primary training objective for an AE is minimizing the discrepancy between input data and reconstructed output, formally expressed as follows:
\begin{equation}
L_{AE} = L_{MSE}(x, D(E(x)),
\end{equation}
where, $x$ denotes input data and $h_z = E(x)$ represents latent embedding.
VAE enhances representation learning by employing variational inference, mapping inputs to probability distributions within the latent space rather than singular points. Consequently, VAEs aim not only for accurate reconstruction but also for aligning latent space distributions with a predefined prior distribution (defined with Gaussian with mean and variance).
Wasserstein Autoencoders leverage the Wasserstein distance metric to regularize the encoded training distribution, aligning it closely with a chosen prior. In WAE-GAN implementations, training involves an adversarial setup between the encoder-decoder model and a discriminator. The training consists of two iterative stages. Initially, both encoder and decoder remain fixed while latent spaces generated by the encoder are evaluated by the discriminator classifying latent codes as original or synthetic. 
The discriminator learns to distinguish between original data and artificially generated synthetic data in the first stage. In the second stage, the model performs for training to improve the data reconstruction ability of the decoder. Thus, the encoder, decoder, and discriminator compete with each other and improve the overall data generation quality through adversarial games that develop in a direction that outperforms the adversarial network.

Our research incorporates feature relationships considering temporal properties to estimate Ground Reaction Forces (GRF) from insole-image videos. We adopt Knowledge Distillation (KD) for efficiency and consider different methods for learning representations in deep learning based estimation.

\subsection{Knowledge Distillation}
Knowledge Distillation (KD) was first proposed by Buciluǎ \emph{et al.} \cite{bucilua2006model}, and later expanded by Hinton \emph{et al.} \cite{hinton2015distilling}. KD presents an effective strategy for developing compact neural networks by transferring knowledge from a larger, pre-trained teacher model to a smaller student model. KD typically employs softened probability outputs or soft labels generated through temperature-scaled logits, which effectively mitigate difference between the teacher and student while providing guidance for learning to the student model. The standard KD loss is formulated as:
\begin{equation}
L_{KD} = \alpha L_{CE} + (1 - \alpha) L_D, 
\end{equation} 
where, $L_{CE}$ represents the cross-entropy loss against ground truth labels, $L_D$ is the KD loss, and hyperparameter $\alpha$ controls their relative contribution. The cross-entropy loss component is defined as:
\begin{equation}
L_{CE} = H(\sigma(y_S), y_g), 
\end{equation} 
where, $\sigma(\cdot)$ denotes the softmax function, $H(\cdot)$ represents cross-entropy loss function, $y_S$ is logit of the student, and $y_g$ is the ground truth. The distillation loss is calculated to minimize divergence between softened outputs of teacher and student networks:
\begin{equation}
L_D = \tau^2 L_{KL}(r_T, r_S),
\end{equation} 
where $\tau > 1$ is a hyperparameter controlling teacher and student output softening, and $r_T=\sigma(y_T/\tau)$ and $r_S=\sigma(y_S/\tau)$ represent softened probabilities from teacher and student, respectively.

To provide additional knowledge in KD learning process, intermediate features from hidden network layers are frequently utilized to strengthen the distillation process \cite{gou2021knowledge, zagoruyko2016paying, tung2019similarity}. Activation-based Attention Transfer (AT) \cite{zagoruyko2016paying} calculates attention maps using channel-wise statistics to guide the learning of the student network effectively. Similarly, Similarity-Preserving knowledge distillation (SP) \cite{tung2019similarity} captures relational knowledge by aligning similarity metrics between teacher and student feature representations within mini-batches, computed as:
\begin{equation} \label{sp_eq}
M = \widehat{G} \cdot \widehat{G}^\top; \quad \widehat{G} \in \mathbb{R}^{b \times chw},
\end{equation}
where the similarity map $M \in \mathbb{R}^{b \times b}$, $\widehat{G}$ represents the intermediate layer's reshaped features, $b$, $c$, $h$, and $w$ indicate mini-batch size, channels, height, and width dimensions, respectively.
With this, the similarity implies correlation of samples within a mini-batch. However, computing relationships of elements in high-dimensional data is a challenge \cite{li2022interpretable, rauker2023toward, johnstone2009statistical, fan2014challenges}. In high dimensions, random correlations tend to appear, and many features that are irrelevant to represent meaningful information. Additionally, computational cost to calculate correlations increases and it makes it challenging to gain effective statistical interpretation of data, often referred to as the `curse of high-dimensionality'. In this light, we propose a framework incorporating sparse correlation with selective features in knowledge distillation, which is effective in transferring knowledge from high-dimensional data.

\section{System Architecture and Proposed Method} \label{sec:proposed_method}
The primary objective of our proposed approach is to develop a framework capable of creating a lightweight model that accurately predicts ground reaction forces (GRF) using an insole sensor in real-world walking environments. To validate our method, we constructed hardware systems for simultaneous data acquisition using both wearable insole sensors and an instrumented split-belt treadmill. We outline the system architecture and introduce our proposed deep learning-based lightweight model generation method below.

\subsection{Hardware Configuration} \label{harset}

\begin{figure*}[htb!]
\includegraphics[width=0.99\linewidth]{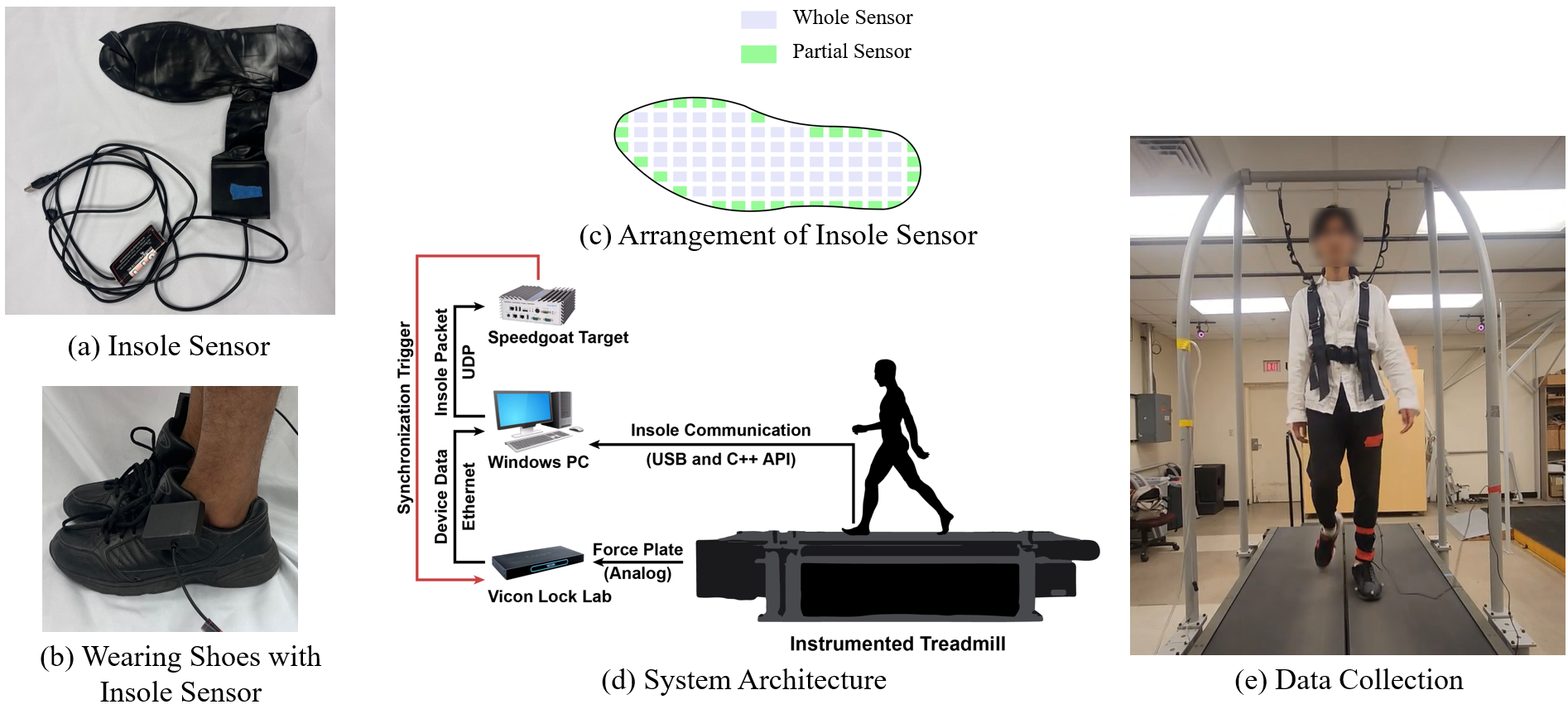} 
\centering
\caption{Description for insole sensors, system architecture, and data collection. Insole sensor is organized with whole and partial sensors. We measured walking behavior on a treadmill while wearing an insole sensor, and collected insole data and GRF data simultaneously.}
\label{figure:data_collection}
\end{figure*}

The sensors used in this study are shown in Fig. \ref{figure:data_collection}. Specifically, we employed the Tactilus V-Series High Performance sensor, developed by SensorProd Inc. (USA), which operates via USB connectivity. This sensor consists of a 16 $\times$ 8 array of piezo-resistive elements arranged in a configuration of 254.0 $\times$ 92.1 $mm$. The insole sensor is approximately the size of a standard shoe, allowing subjects to wear it comfortably, as illustrated in Fig. \ref{figure:data_collection}(b). However, the insoles include partial sensors that can result in noisy or missing data. The sensor is capable of measuring pressure up to 206.8 $kPa$, with an accuracy of $\pm$10\%, repeatability of $\pm$2\%, hysteresis of $\pm$5\%, and non-linearity of $\pm$1.5\%. For sensor data communication, we used a manufacturer-supplied API implemented as a C++ library on a Windows-based system. 
As depicted in the system architecture in Fig. \ref{figure:data_collection}(d), real-time data collection was performed using a multi-threaded C++ application, where one thread continuously updates the current sensor frame, while another simultaneously transmits the data via User Datagram Protocol (UDP) to a Speedgoat Baseline Target machine (Speedgoat GmbH, Switzerland). This parallel processing minimizes sensor latency and enables data acquisition at around 200 $Hz$.
For gait data collection, we utilized an instrumented split-belt treadmill (Bertec Inc., USA), which capture force plate data using the Vicon system (Vicon Motion Systems Ltd., UK) at a frequency of 2000 $Hz$. 
System synchronization was achieved by remotely triggering the Vicon system through the Speedgoat machine.

\subsection{Data Collection}
As described in Fig. \ref{figure:data_collection}(e), we collected data using both the insole sensor and the instrumented treadmill for GRF estimation. Eight healthy participants (six males, two females; age: 29$\pm$5 years; height: 174$\pm$8 $cm$; weight: 64$\pm$6 $kg$) participated in treadmill-based gait data collection at various walking speeds. Due to the limited insole size, all participants wore size 9.5M footwear. Participants were informed about the experimental procedures and provided written consent in accordance with protocols approved by the Arizona State University Institutional Review Board (STUDY00014244).

Each participant walked on a treadmill for 10-minutes during four separate sessions, each at a predefined speed and with a 0$^\circ$ incline. The walking speeds were: slow (SW, 0.88 $m/s$), regular (RW, 1 $m/s$), brisk (BW, 1.25 $m/s$), and fast (FW, 1.5 $m/s$). For each session, the first 10 seconds were allocated for participants to adjust to the speed; thus, data from this period were excluded from the analysis. To reduce sequence effects, the order of sessions was randomized. In addition, mandatory 10-minute rest intervals were provided between sessions to prevent participant fatigue and avoid sensor saturation.

\begin{figure}[htb!]
\includegraphics[scale=0.34] {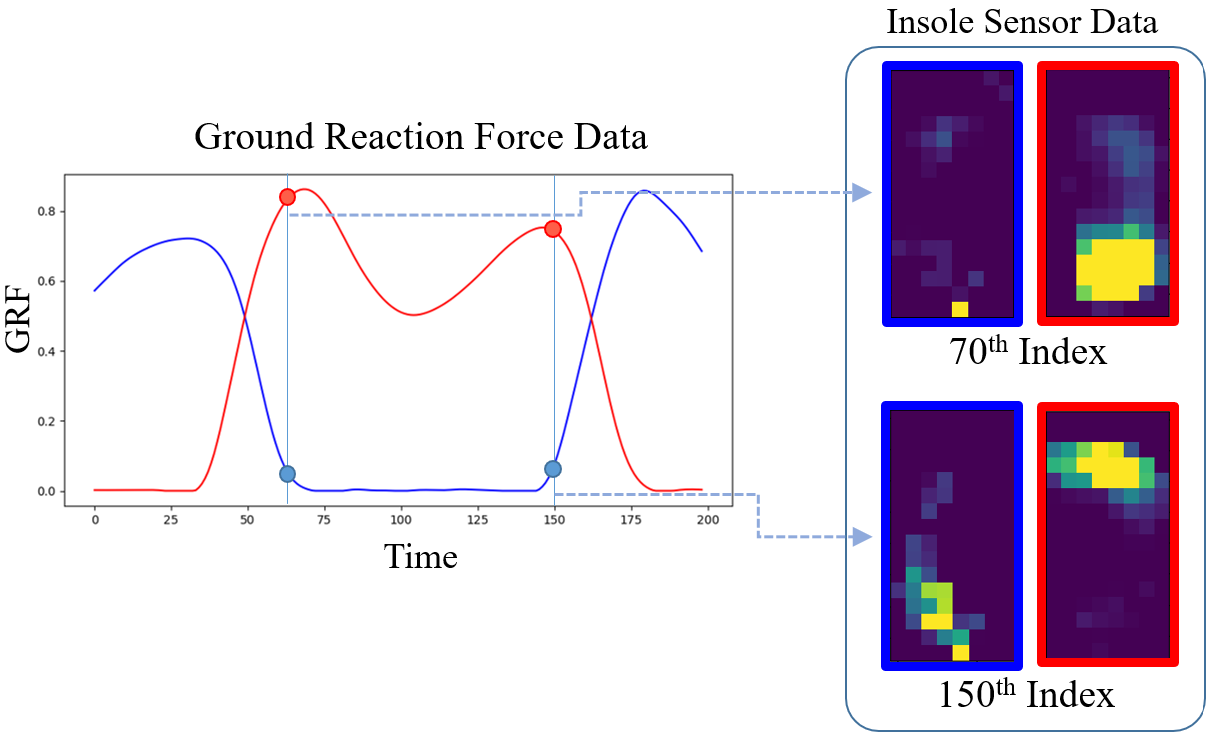} 
\centering
\caption{An example of ground reaction force (GRF) data and its corresponding insole sensor data. Blue and red colored lines in the graph present GRF data for the left and right foot, respectively. The blue and red boxes display the insole sensor data for their corresponding GRF data.}
\label{figure:grf_insole}
\end{figure}

Ground reaction force (GRF) data obtained from the force plate were downsampled to 200 $Hz$ to match the sampling rate of the insole sensor. The GRF signals were then processed using a second-order, zero-lag Butterworth low-pass filter with a 10 $Hz$ cutoff frequency. For insole data, spatial filtering was applied to the pressure matrices (16$\times$8$\times t$, where $t$ represents the time frames) using a fraction matrix that accounts for effective pixel coverage. Pixels fully enclosed within the insole boundary were assigned a value of 1, while edge pixels were assigned values of 0.33 or 0.67 based on visual estimation of their area relative to a full pixel. Pixels located outside the boundary or missing due to the shape constraints of the insole were assigned a value of 0. This fraction matrix was applied element-wise to the insole pressure data to reduce errors caused by load concentrations near the edges, which are commonly attributed to deformation of the flexible insole. Fig. \ref{figure:grf_insole} shows the GRF data paired with corresponding insole sensor data, with blue and red lines representing the left and right feet, respectively.

\subsection{Proposed Method}
\subsubsection{Spatiotemporal Networks}
To generate outputs in the form of time-series data representing GRFs from insole sensor data with sequential frames, we adopt encoder-decoder networks. Since the input data includes time step information in the form of sequential images, it can be considered as a video. This allows to capture the temporal properties of the data, providing richer and more informative features compared to using a single static image.

To extract meaningful information from the insole sensor data, we employ various spatiotemporal convolutional networks that can encode both spatial information at each time step and temporal variances across time. Our encoder networks are constructed using different architectures, including 3D ConvNets (C3D) \cite{tran2015learning}, Inflated 3D ConvNets (I3D) \cite{carreira2017quo}, and (2+1)D networks \cite{tran2018closer}, which utilize different types of convolutional kernels and layer configurations. These architectures enable flexible construction of teacher-student combinations for knowledge distillation (KD) to evaluate the performance of diverse approaches. The filters in these networks capture both temporal and spatial features simultaneously.
Specifically, C3D uses 3D convolution layers, I3D introduces a two-stream structure by inflating 2D convolution and pooling kernels, and (2+1)D decomposes 3D convolutions into sequential operations of 2D spatial and 1D temporal convolutions


To reconstruct time-series data from the output of the encoder, the decoder network is composed of 1D CNNs, which help in capturing temporal properties and aligning the output dimensions with the time-series ground truth data. Additionally, the encoded representation can be easily adopted as input to this network architecture. Various types and combinations of encoder-decoder networks can be designed. The specific models used in our experiments are described in Section \ref{network_architecture}.

\subsubsection{Knowledge Relationship Transfer}
We utilize knowledge distillation, a promising method for generating a smaller model by leveraging the knowledge of a larger model. For effective knowledge transfer, we use outputs from intermediate layers, the decoder, and the final results to guide the student model in mimicking the performance of the teacher model through supervised learning. Firstly, we employ a mean squared error (MSE) loss to encourage the student model to generate outputs that closely match the ground truth obtained from the treadmill. The loss function is defined as follows:
\begin{equation}
    \mathcal{L}_{gt} = \mathcal{L}_{MSE}(y_{gt},D(E(x))),
\end{equation}
where, $E$ and $D$ are encoder and decoder networks, $x$ and $y_{gt}$ are input data (insole sensor data) and ground truth (GRF from treadmill), respectively.

Building on the principles of conventional knowledge distillation (KD), which typically involves matching the logits of teacher and student models, we utilize the outputs of the decoder networks from both the teacher and student. To encourage the student model to mimic the estimation performance of teacher, we construct a loss function based on cosine similarity, which reflects the Pearson correlation coefficient~\cite{pearson} under a stochastic distribution.
In our approach, we consider probabilistic features for both inter- and intra-temporal extents during distillation, which implies relationships within the same time intervals and temporal trends over time for estimated sequences. These additional relationships within the estimated results of the teacher provide richer guidance for the student model. The loss function used for knowledge transfer, incorporating both inter- and intra-relations, is defined as follows:
\begin{equation}
\label{eq_intra}
\begin{split}
    \mathcal{L_{KD}}_{inter} &= 1 - \frac{1}{b}\sum_{i=1}^{b}\rho_p(h_{{T[i,:]}}, h_{{S[i,:]}}), \\
    \mathcal{L_{KD}}_{intra} &= 1 - \frac{1}{t}\sum_{j=1}^{t}\rho_p(h^{\top}_{{T[j,:]}}, h^{\top}_{{S[j,:]}}), \\
    \mathcal{L_{KD}}_{c} &= \frac{1}{c}\sum_{\text{i}=1}^{c}(\mathcal{L_{KD}^{\text{i}}}_{inter} + \mathcal{L_{KD}^{\text{i}}}_{intra}),
\end{split}   
\end{equation}
where, $h$ is the results for softmax of the predicted output from the decoder, $\mathcal{L_{KD}}_{inter}$ and $\mathcal{L_{KD}}_{intra}$ are losses for inter- and intra-temporal extents, respectively, $\rho_p(\cdot, \cdot)$ is the Pearson correlation coefficient between two variables~\cite{pearson}, and $t$ is temporal extent (window length).





In the proposed method, we not only consider the predicted outputs of the teacher and student models but also leverage knowledge from intermediate layers. To facilitate effective knowledge transfer, we utilize relationships within a mini-batch, representing feature similarity \cite{tung2019similarity}, as described in \eqref{sp_eq}.
Unlike algorithms that analyze a single image at a specific point in time, preserving temporal features is essential when implementing with sequential data. In this context, we introduce a method that leverages correlation across selective channels for knowledge distillation.
We firstly explain utilizing the output of the encoder, which represents compressed features at the midpoint of the network. To better align with the characteristics of time-series data, our method emphasizes preserving both feature similarity and temporal properties within a mini-batch. Computing with high-dimensional representations is essential when working with spatiotemporal data. However, as dimensionality increases, extracting meaningful relationships among individual elements and interpreting their characteristics becomes increasingly challenging \cite{rauker2023toward, johnstone2009statistical, fan2014challenges}. To address this issue, we devise an effective method to capture relational features within feature elements. We extract correlation of features selected from specific channels over chosen intervals rather than using whole features, considering less complexity as well as improvement of KD in training process. A mapping function to obtain the correlation matrix, $G$, is introduced as follows:
\begin{equation}\label{eq_corr}
 \psi: F \rightarrow G \in \mathbb{R}^{b\times b},
\end{equation}
where, $F \in \mathbb{R}^{b\times t}$ is a feature chosen at an index of channels from an encoded feature, $\widetilde{F} \in \mathbb{R}^{b\times c \times t}$, which indicates features from intermediate layers or output of the encoder. We provide more details in the next section. Any correlation metric can be used to extract a correlation matrix. We utilize Gaussian RBF kernel function that performs better than na\"ive MMD and Bilinear Pool for capturing in capturing the complex non-linear relationship between instances \cite{scholkopf1997comparing, ring2016approximation}. The function $K$ can be written as follows:
\begin{equation}\label{eq_rbf}
\begin{split}
 [K(F, F)]_{ij} = exp(-\gamma\norm{F_i - F_j}^2) \\
 \approx \sum_{p=0}^{P}  exp(-2\gamma) \frac{(2\gamma)^p}{p!}(F_i \cdot F_j^\top)^p,
\end{split}
\end{equation}
where the pairwise correlations between $i^{th}$ and $j^{th}$ features in $F$ are encoded as elements of $[K(F,F)]_{ij}$, and each element of $[K(F,F)]_{ij}$ can be approximated by a $P^{th}$-order Taylor series.
The knowledge transfer with extracted correlation with channels over chosen intervals is computed as follows:
\begin{equation}\label{eq_cmap}
\begin{split}
    \mathcal{L}_{sc} &= \frac{1}{b^2q|L|}
    \sum_{(l^T, l^S) \in L} \biggl( \sum_{k \in c^T} \bignorm{ \psi(F^{l^T}_k) - \psi(F^{l^S}_k) }^{2}_{2} \biggl) \\
    &= \frac{1}{b^2q|L|}
    \sum_{(l^T, l^S) \in L} \biggl( \sum_{k \in c^T} \bignorm{ G^{l^T}_{k} - G^{l^S}_{k} }^{2}_{2} \biggl),
\end{split}
\end{equation}
where $G^{l^T}$ and $G^{l^S}$ are correlations maps implying similarity of features for layer pairs ($l^T$ and $l^S$) from a teacher and a student, respectively, $L$ collects the layer pairs, and $k$ denotes an index of a channel determined by chosen intervals. Specifically,
$k$ ranges from 0 to $c$ and is selectively chosen at regular intervals of $m \geq 1$.
For example, if $c = rm + n$ where $0 \leq n < m$, then $k \in \{0, m, 2m, \dots, rm\}$.
Here, $r$ is the integer quotient of $c$ divided by $m$, i.e., $r = \lfloor c / m \rfloor$. Consequently, $q$ is the number of selected correlation maps, defined by $r+1$.



\subsubsection{Knowledge Transfer with Intermediate Layers}
In knowledge transfer, we utilize diverse features from different layers in a model including encoder and decoder networks. We use compressed representation that is middle representation -- output of an encoder -- as a default knowledge used for knowledge distillation. 


Additionally, features from other intermediate layers of the encoder and decoder can be leveraged during knowledge transfer to improve the training performance of the student model, as high- and mid-level layers contribute differently to the learning process \cite{jeong2022lightweight}. In the proposed method, we utilize the middle representation along with features from additional layers of the encoder and decoder that are near the middle layer.
When extracting features from the encoder, the feature $\widehat{F}_{e} \in \mathbb{R}^{b\times c \times t \times w \times h}$ includes spatial information represented by $w$ and $h$.
To focus on temporal knowledge and its correlations, spatial information is aggregated through average pooling, resulting in $\widetilde{F}_{e}\in \mathbb{R}^{b\times c \times t}$. Feature, $\widetilde{F}$, used in the extraction of correlation, preserves temporal extent, $t$, which is not reduced even if spatial information is encoded and specific channels are chosen in the distillation process. This preservation of temporal features is important for better generation of time-series data as output. Since the architectures of the teacher and student networks may differ, the temporal dimensions of their features can also vary. In cases of dimensional mismatch, bilinear interpolation is applied to align the temporal dimensions of the teacher and student feature maps. We investigate the effects of utilizing features from various layers in Section \ref{layer_selec}.

\subsubsection{Final Objective of SCKD}

The learning objective of the proposed method is: 
\begin{equation}
    \mathcal{L}_\text{SCKD} = \mathcal{L}_{gt} + \lambda_1\mathcal{L}_{KD_c} + \lambda_2\mathcal{L}_{{sc}_{r}} + \lambda_3\mathcal{L}_{{sc}_{f}},
\end{equation}
where, $\mathcal{L}_{{sc}_{r}}$ and $\mathcal{L}_{{sc}_{f}}$ are loss values for compressed representations from the middle layer and intermediate features from layers in encoder or decoder networks, respectively. $\lambda_1$ is a hyperparameter to control the effect of estimated result comparison between the teacher and student, $\lambda_2$ and $\lambda_3$ are hyperparameters to balance effects between compressed representations and intermediate features of encoder and decoder networks.



\section{Experimental Results and Analysis} \label{sec:experiments}

In this section, we describe the datasets and settings for our experiments. The proposed framework is evaluated using datasets with varying window lengths and through various combinations of teacher-student networks consisting of different learning methods.

\subsection{Experimental Settings} \label{sec:settings}

\subsubsection{Data Description}
For sequential data analysis, the choice of window length affects performance.
To evaluate the impact of window length, we construct datasets using fully non-overlapping sliding windows of varying lengths.

\begin{table*}[htb!]
\renewcommand{\tabcolsep}{1.2mm} 
\caption{Details of dataset consisting of different settings with various speeds recorded for eight subjects.}
\label{table:Dataset}

\begin{center}

\scalebox{0.9}{
\begin{tabular}{c|c| c c c c c c c c |c c}

\hline
\centering
Window & Walking speed & Sbj.01 & Sbj.02 &
Sbj.03 & Sbj.04 & Sbj.05 & Sbj.06 & 
Sbj.07 & Sbj.08 & Sum & \# of subjects\\
\hline
 \multirow{4}{*}{200}  & SW (0.88 $m/s$) & 279 & 279 & 279 & 279 & 279 & 279 & 279 & 279 & 2232 & 8\\ 
   & RW (1.00 $m/s$) & 0 & 0 & 279 & 279 & 279 & 279 & 279 & 279 & 1674 & 6\\ 
   & BW (1.25 $m/s$) & 0 & 279 & 279 & 279 & 279 & 279 & 279 & 279 & 1953 & 7\\ 
   & FW (1.50 $m/s$) & 279 & 279 & 279 & 0 & 279 & 279 & 279 & 279 & 1953 & 7\\ \hline
 \multirow{4}{*}{100}  & SW (0.88 $m/s$) & 559 & 559 & 559 & 559 & 559 & 559 & 559 & 559 & 4472 & 8\\ 
   & RW (1.00 $m/s$) & 0 & 0 & 559 & 559 & 559 & 559 & 559 & 559 & 3354 & 6\\ 
   & BW (1.25 $m/s$) & 0 & 559 & 559 & 559 & 559 & 559 & 559 & 559 & 3913 & 7\\ 
   & FW (1.50 $m/s$) & 559 & 559 & 559 & 0 & 559 & 559 & 559 & 559 & 3913 & 7\\
\hline

\end{tabular}
}
\end{center}

\end{table*}

In the collected data, a single gait cycle is typically represented within 200 time steps (window size). Based on this observation, we set the default window size to 200 for the following experiments. To evaluate the methods under different conditions, we also include a smaller window size of 100, which imposes additional challenges and may better highlight performance differences among various methods. Table \ref{table:Dataset} provides details, including statistics and conditions of the datasets. Walking speed is expressed in meters per second. The distribution of samples across the dataset is uneven for all window sizes. Since no additional pre- or post-processing is applied, the samples consist of raw data that may include perturbations and unknown noise, increasing the difficulty of analysis. The data is evaluated using a leave-one-subject-out cross-validation approach, with non-overlapped data samples to consider the effects of individual variability. GRF data is normalized by each subject’s body weight (expressed as a percentage of body weight) as measured by the instrumented treadmill. The data is further normalized to the range $[0, 1]$. The corresponding insole data (\textit{n}-pixels measuring 0-30 \textit{psi} each) is also normalized to the range $[0, 1]$, thereby both data can be easily utilized in model learning.
Since we train a model with data from both feet, the size of channels for both data for the treadmill and insole is 2.


\subsubsection{Network Architectures} \label{network_architecture}






We construct teacher and student models using different network architectures with C3D, I3D, and (2+1)D structures as encoders, which are widely adopted for sequential data (e.g., video and temporal frames) analysis \cite{ji20123d, huang2023review}.
Table \ref{table:info_settings} presents various combinations of teacher and student networks. The teacher models consist of C3D, I3D, and (2+1)D encoders, denoted as Teacher1, Teacher2, and Teacher3, respectively. These teachers offer diverse effects in the distillation process used to train the student model.
The table also details the floating point operations per second (FLOPs), the number of trainable parameters, and the model compression ratio. FLOPs is a measure of computational performance in implementation. These metrics highlight the varying specifications of each network.
In our experiments, the student model has significantly fewer parameters, indicating that it requires substantially fewer operations compared to the teacher models.

\begin{table}[htb!]
\centering
\renewcommand{\tabcolsep}{0.9mm} 
\caption{Details of teacher-student networks. Brackets denote the type of encoder network architecture. $\dagger$ denotes small size of network representing a student. T and S indicate a teacher and student, respectively. The compression ratio (Comp. ratio) is calculated with the number of parameters of student to teacher.}

\begin{center}
\begin{tabular}{c |c |c c c |c c c}

\hline
\multicolumn{2}{c|}{Window} & \multicolumn{3}{c|}{200} & \multicolumn{3}{c}{100} \\ \hline


\multicolumn{2}{c|}{\multirow{2}{*}{Teacher}} & Teacher1 & Teacher2 & Teacher3 & Teacher1 & Teacher2 & Teacher3 \\ 
\multicolumn{2}{c|}{} & (C3D) & (I3D) & ((2+1)D) & (C3D) & (I3D) & ((2+1)D) \\ \hline

\multicolumn{2}{c|}{Student} & \multicolumn{3}{c|}{C3D$^\dagger$} & \multicolumn{3}{c}{C3D$^\dagger$} \\
\hline

FLOPs & T & 635.39 & 745.90 & 1580.11 & 308.66 & 373.51 & 790.61 \\  
(M) & S & \multicolumn{3}{c|}{164.55} & \multicolumn{3}{c}{79.98} \\ \hline

\#params & T & 1.00 & 1.18 & 1.58 & 1.00 & 1.18& 1.58 \\ 
(M) & S & \multicolumn{3}{c|}{0.25} & \multicolumn{3}{c}{0.25} \\ \hline

\multicolumn{2}{c|}{Comp. ratio} & 25.10$\%$ & 21.35$\%$ & 15.88$\%$ & 25.10$\%$ & 21.35$\%$ & 15.88$\%$ \\
\hline

\end{tabular}
\end{center}
\label{table:info_settings}
\end{table}


\subsubsection{Training and Evaluation}
For model training, we set the total number of epochs to 200 with a batch size of 128. The Adam optimizer is used with a learning rate of 0.01 for networks incorporating C3D and I3D encoders, and 0.001 for those using the (2+1)D encoder. Hyperparameters for the loss functions are set empirically. Specifically, we set $\lambda_{1}$, $\lambda_{2}$, and $\lambda_{3}$ to 1, 10, and 1, respectively, and assign values of 0.4, 2, and 8 to $\gamma$, $P$, and $q$, respectively.

Following previous studies \cite{orlin2000plantar}, we evaluate model performance using root mean squared error (RMSE, $\times10^{-2}$), mean absolute error (MAE, $\times10^{-2}$), and the Pearson correlation coefficient ($r$, $\times10^{-2}$) between predicted data from the insole sensor and ground truth measurements from the treadmill. 
RMSE and MAE are complementary metrics to evaluate model performance.
More specifically, RMSE emphasizes outliers and poor predictions, which helps interpret if the model may generate outliers that are undesirable.
The MAE measures average absolute difference between predicted one from insole data and true GRF values by treadmill, which helps easy interpretation of errors and general prediction.
The correlation coefficient reflects the ratio of the covariance between two variables to the product of their standard deviations. That is, a value close to 1 indicates a strong correlation between the two data sources. Each experiment is repeated three times, and we report the average results along with the standard deviation.

To demonstrate model generalizability, we adopt a leave-one-subject-out evaluation metric. For baseline comparisons, we include conventional knowledge distillation (KD) \cite{hinton2015distilling}, attention transfer (AT) \cite{zagoruyko2016paying}, similarity-preserving knowledge distillation (SP) \cite{tung2019similarity}, DIST \cite{huang2022knowledge}, and semantic calibration for cross-layer knowledge distillation (SemCKD) \cite{semckd}. In applying these baselines to our estimation task, we utilize the intermediate representations for knowledge transfer. For DIST, the decoder output is used in accordance with its original approach. The KD hyperparameter $\lambda$ is empirically set to 1 for both DIST and SemCKD. All other baselines are configured using the same settings as our method.

\subsection{Analysis on Network Architectures} \label{sec.4_2}
To analyze the effectiveness of the proposed method, we investigate performance on different architectural combinations of teacher and student networks. Also, we evaluate the performance using various lengths of input sequences, specifically 100 and 200 time steps. The teacher models use encoders based on C3D, I3D, and (2+1)D architectures, denoted as Teacher1, Teacher2, and Teacher3, respectively. The student model is constructed with a C3D encoder of significantly smaller capacity compared to the teacher models. In this section, both teacher and student models are trained using an autoencoder-based learning approach. Model performance is evaluated for ground reaction force (GRF) estimation using RMSE, MAE, and Pearson correlation coefficient ($r$).

Table \ref{table:grf_w100w200} reports the results for the vanilla model as well as various knowledge distillation methods. The results show that the proposed method, SCKD, achieves the best performance across all metrics, in most cases, characterized by lower RMSE and MAE, and higher $r$ values exceeding 90\%. Notably, the proposed method with Teacher1 outperforms even the teacher models for both window lengths, highlighting the significant effectiveness of our approach in generating improved student models through knowledge distillation.
On the other hand, baseline methods such as DIST and SemCKD require modifications to the network architecture for loss computation and rely on additional hidden layers for knowledge transfer, which increases computational overhead. Furthermore, when the network structures of teacher and student models differ substantially (e.g., in the case of Teacher3), these baseline methods show degraded performance. In some cases, these methods show even worse performance than that of the vanilla student model. This tendency is particularly pronounced with the shorter input window length of 100 time steps. In general, all methods perform better with a window length of 200 compared to 100, indicating that longer sequences provide more informative features, benefiting both knowledge distillation and GRF estimation.


\begin{table}[htb!]
\centering
\caption{Comparison of RMSE, MAE, and $r$ with student model distilled by various methods on ground reaction force estimation. \textbf{Bold} denotes best results for a distilled student.}

\begin{center}
\scalebox{0.85}{
\begin{tabular}{c |c c c |c c c}

\hline
\multirow{2}{*}{Method} & \multicolumn{3}{c|}{W200} & \multicolumn{3}{c}{W100} \\ \cline{2-7}



& RMSE$\downarrow$ & MAE$\downarrow$ & $r\uparrow$ & RMSE$\downarrow$ & MAE$\downarrow$ & $r\uparrow$ \\ \hline

\multirow{2}{*}{Student} & 6.253 & 4.558 & 98.757 & 6.458 & 4.731 & 94.246 \\
& {\scriptsize$\pm$0.259} & {\scriptsize$\pm$0.162} & {\scriptsize$\pm$0.071} & {\scriptsize$\pm$0.229} &
{\scriptsize$\pm$0.158} & {\scriptsize$\pm$0.390} \\ \hline \hline

\multirow{2}{*}{Teacher1} & 5.978 & 4.376 & 98.740 & 6.321 & 4.648 & 94.552 \\
& {\scriptsize$\pm$0.189} & {\scriptsize$\pm$0.117} & {\scriptsize$\pm$0.073} & {\scriptsize$\pm$0.216} &
{\scriptsize$\pm$0.154} & {\scriptsize$\pm$0.334} \\ \hline

\multirow{2}{*}{KD} & 6.264 & 4.577 & 98.678 & 6.476 & 4.764 & 94.466 \\
& {\scriptsize$\pm$0.248} & {\scriptsize$\pm$0.159} & {\scriptsize$\pm$0.089} & {\scriptsize$\pm$0.209} &
{\scriptsize$\pm$0.145} & {\scriptsize$\pm$0.390} \\

\multirow{2}{*}{AT} & 6.232 & 4.550 & 98.708 & 6.507 & 4.785 & 94.293 \\
& {\scriptsize$\pm$0.236} & {\scriptsize$\pm$0.150} & {\scriptsize$\pm$0.078} & {\scriptsize$\pm$0.235} &
{\scriptsize$\pm$0.165} & {\scriptsize$\pm$0.390} \\


\multirow{2}{*}{SP} & 6.228 & 4.560 & 98.705 & 6.482 & 4.742 & 94.618 \\
& {\scriptsize$\pm$0.233} & {\scriptsize$\pm$0.151} & {\scriptsize$\pm$0.078} & {\scriptsize$\pm$0.237} &
{\scriptsize$\pm$0.165} & {\scriptsize$\pm$0.416} \\

\multirow{2}{*}{KD+SP} & 6.248 & 4.582 & 98.707 & 6.465 & 4.726 & 94.379 \\
& {\scriptsize$\pm$0.237} & {\scriptsize$\pm$0.153} & {\scriptsize$\pm$0.080} & {\scriptsize$\pm$0.238} &
{\scriptsize$\pm$0.168} & {\scriptsize$\pm$0.421} \\

\multirow{2}{*}{DIST} & 6.106 & 4.495 & 98.779 & 6.260 & 4.627 & \textbf{94.879} \\
& {\scriptsize$\pm$0.214} & {\scriptsize$\pm$0.128} & {\scriptsize$\pm$0.075} & {\scriptsize$\pm$0.141} &
{\scriptsize$\pm$0.085} & {\scriptsize$\pm$0.336} \\

\multirow{2}{*}{SemCKD} & 6.221 & 4.524 & 98.631 & 6.400 & 4.719 & 93.749 \\
& {\scriptsize$\pm$0.181} & {\scriptsize$\pm$0.112} & {\scriptsize$\pm$0.070} & {\scriptsize$\pm$0.163} &
{\scriptsize$\pm$0.112} & {\scriptsize$\pm$0.433} \\ 

\hline



\multirow{2}{*}{SCKD} & \textbf{5.941} & \textbf{4.351} & \textbf{98.788} & \textbf{6.100} & \textbf{4.522} & 94.531 \\
& {\scriptsize$\pm$0.217} & {\scriptsize$\pm$0.137} & {\scriptsize$\pm$0.077} & {\scriptsize$\pm$0.151} &
{\scriptsize$\pm$0.092} & {\scriptsize$\pm$0.396} \\ \hline \hline


\multirow{2}{*}{Teacher2} & 5.382 & 3.962 & 99.027 & 5.520 & 4.056 & 96.306 \\
& {\scriptsize$\pm$0.226} & {\scriptsize$\pm$0.163} & {\scriptsize$\pm$0.084} & {\scriptsize$\pm$0.252} &
{\scriptsize$\pm$0.185} & {\scriptsize$\pm$0.280} \\ \hline

\multirow{2}{*}{KD} & 6.293 & 4.604 & 98.694 & 6.430 & 4.723 & 94.369 \\
& {\scriptsize$\pm$0.277} & {\scriptsize$\pm$0.184} & {\scriptsize$\pm$0.081} & {\scriptsize$\pm$0.233} &
{\scriptsize$\pm$0.168} & {\scriptsize$\pm$0.388} \\

\multirow{2}{*}{AT} & 6.268 & 4.543 & 98.652 & 6.460 & 4.751 & 94.475 \\
& {\scriptsize$\pm$0.268} & {\scriptsize$\pm$0.167} & {\scriptsize$\pm$0.102} & {\scriptsize$\pm$0.227} &
{\scriptsize$\pm$0.163} & {\scriptsize$\pm$0.353} \\

\multirow{2}{*}{SP} & 6.297 & 4.600 & 98.668 & 6.460 & 4.723 & 94.369 \\
& {\scriptsize$\pm$0.263} & {\scriptsize$\pm$0.174} & {\scriptsize$\pm$0.087} & {\scriptsize$\pm$0.230} &
{\scriptsize$\pm$0.164} & {\scriptsize$\pm$0.396} \\

\multirow{2}{*}{KD+SP} & 6.304 & 4.615 & 98.674 & 6.442 & 4.738 & 94.341 \\
& {\scriptsize$\pm$0.246} & {\scriptsize$\pm$0.161} & {\scriptsize$\pm$0.078} & {\scriptsize$\pm$0.231} &
{\scriptsize$\pm$0.164} & {\scriptsize$\pm$0.390} \\

\multirow{2}{*}{DIST} & 6.098 & 4.406 & 98.789 & 6.398 & 4.698 & 94.517 \\
& {\scriptsize$\pm$0.172} & {\scriptsize$\pm$0.124} & {\scriptsize$\pm$0.052} & {\scriptsize$\pm$0.212} &
{\scriptsize$\pm$0.164} & {\scriptsize$\pm$0.256} \\

\multirow{2}{*}{SemCKD} & 6.197 & 4.490 & 98.660 & 6.374 & 4.722 & 93.723 \\
& {\scriptsize$\pm$0.196} & {\scriptsize$\pm$0.126} & {\scriptsize$\pm$0.071} & {\scriptsize$\pm$0.198} &
{\scriptsize$\pm$0.137} & {\scriptsize$\pm$0.382} \\

\hline



\multirow{2}{*}{SCKD} & \textbf{5.784} & \textbf{4.299} & \textbf{98.901} & \textbf{6.315} & \textbf{4.624} & \textbf{95.598} \\
& {\scriptsize$\pm$0.158} & {\scriptsize$\pm$0.112} & {\scriptsize$\pm$0.045} & {\scriptsize$\pm$0.172} &
{\scriptsize$\pm$0.140} & {\scriptsize$\pm$0.239} \\ \hline \hline

\multirow{2}{*}{Teacher3} & 5.623 & 4.091 & 98.849 & 5.754 & 4.188 & 95.782 \\
& {\scriptsize$\pm$0.252} & {\scriptsize$\pm$0.163} & {\scriptsize$\pm$0.124} & {\scriptsize$\pm$0.306} &
{\scriptsize$\pm$0.203} & {\scriptsize$\pm$0.473} \\ \hline

\multirow{2}{*}{KD} & 6.284 & 4.584 & \textbf{98.715} & 6.554 & 4.814 & 94.291 \\
& {\scriptsize$\pm$0.271} & {\scriptsize$\pm$0.176} & {\scriptsize$\pm$0.074} & {\scriptsize$\pm$0.241} &
{\scriptsize$\pm$0.166} & {\scriptsize$\pm$0.373} \\

\multirow{2}{*}{AT} & 6.245 & 4.543 & 98.711 & 6.487 & 4.771 & 94.311 \\
& {\scriptsize$\pm$0.266} & {\scriptsize$\pm$0.168} & {\scriptsize$\pm$0.088} & {\scriptsize$\pm$0.235} &
{\scriptsize$\pm$0.166} & {\scriptsize$\pm$0.390} \\

\multirow{2}{*}{SP} & 6.236 & 4.544 & 98.672 & 6.606 & 4.842 & 94.176 \\
 & {\scriptsize$\pm$0.258} & {\scriptsize$\pm$0.168} & {\scriptsize$\pm$0.092} & {\scriptsize$\pm$0.254} &
{\scriptsize$\pm$0.173} & {\scriptsize$\pm$0.395} \\

\multirow{2}{*}{KD+SP} & 6.221 & 4.530 & 98.675 & 6.586 & 4.842 & 94.176 \\
& {\scriptsize$\pm$0.253} & {\scriptsize$\pm$0.163} & {\scriptsize$\pm$0.091} & {\scriptsize$\pm$0.253} &
{\scriptsize$\pm$0.173} & {\scriptsize$\pm$0.396} \\

\multirow{2}{*}{DIST} & 6.245 & 4.537 & 98.682 & 6.525 & 4.748 & 95.084 \\
& {\scriptsize$\pm$0.272} & {\scriptsize$\pm$0.165} & {\scriptsize$\pm$0.106} & {\scriptsize$\pm$0.214} &
{\scriptsize$\pm$0.163} & {\scriptsize$\pm$0.471} \\

\multirow{2}{*}{SemCKD} & 6.315 & 4.574 & 98.548 & 6.483 & 4.778 & 93.583 \\
& {\scriptsize$\pm$0.184} & {\scriptsize$\pm$0.098} & {\scriptsize$\pm$0.083} & {\scriptsize$\pm$0.172} &
{\scriptsize$\pm$0.130} & {\scriptsize$\pm$0.339} \\ 

\hline



\multirow{2}{*}{SCKD} & \textbf{6.141} & \textbf{4.493} & 98.678 & \textbf{6.422} & \textbf{4.690} & \textbf{95.084} \\
& {\scriptsize$\pm$0.266} & {\scriptsize$\pm$0.166} & {\scriptsize$\pm$0.109} & {\scriptsize$\pm$0.219} &
{\scriptsize$\pm$0.148} & {\scriptsize$\pm$0.474} \\

\hline

\end{tabular}
}
\end{center}
\label{table:grf_w100w200}
\end{table}


\begin{figure}[htb!]
\includegraphics[scale=0.35] {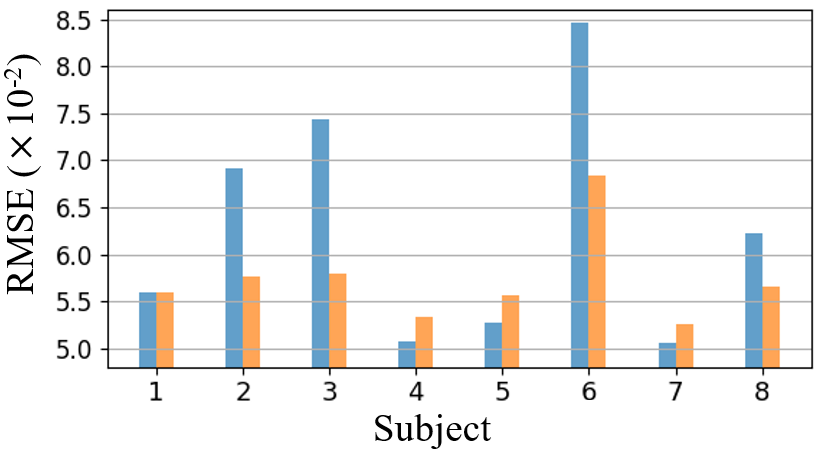} 
\centering
\caption{Results (RMSE) for GRF estimation of Student (blue) and student model distilled by the proposed method (orange), SCKD, using Teacher2 with 200 of window size across subjects. In Student model, large variations in the estimated GRF during walking can occur across individuals; the proposed method, SCKD, mitigates this issue and provides more accurate estimations.}
\label{figure:sbj_result}
\end{figure}

Additionally, we show results of RMSE for Student (learned from scratch) and a student distilled by SCKD with Teacher2 on 200 window size across subjects. As shown in Fig. \ref{figure:sbj_result}, SCKD outperforms Student in overall cases. For subject 6, RMSE of Student is much larger than other cases, which implies that this subject has less common patterns compared to others. This shows cutomizable system is required to handle this personal variability issues and the model fine-tuning or further re-training processes are required.
Even with this case, SCKD performs better than Student, presenting robustness on personal variability.
Importantly, SCKD achieves much more stable results across different subjects. For instance, the results from Student vary on subjects compared to the ones from SCKD. This effectively alleviates the further fine-tuning burdens. With this observation, we found that the model distilled by SCKD performs stable, and is less sensitive and affected by personal variability and perturbations.

\subsection{Analysis on Training Approaches}
To investigate training methods for encoder and decoder networks, we utilize teacher models trained with AE \cite{kingma2013auto}, VAE \cite{fabius2015variational}, and WAE \cite{tolstikhin2018wasserstein} in the distillation process. For WAE, the discriminator is constructed with five linear and four ReLU layers, which contains approximately 155 million trainable parameters. When training a teacher model with VAE, the resulting RMSE is significantly higher than that of both AE and WAE. Prior studies have suggested that the variational nature of VAE may not be ideal for tasks that involves the analysis of underlying data structures \cite{sejnova2023benchmarking}. Specifically, the objective of VAE to estimate the latent distribution may not be suitable for the characteristics of the data at hand \cite{daunhawerlimitations}.
To focus on more stable configurations, in this section we investigate the effectiveness of teacher models based on AE and WAE.

\begin{table}[htb!]
\centering
\caption{Results of RMSE, MAE, and $r$ for GRF estimation of student model distilled by SCKD on WAE teachers. Brackets denote improvement compared to AE.}

\renewcommand{\tabcolsep}{1.3mm} 

\begin{center}
\scalebox{0.95}{
\begin{tabular}{ c |c c c |c c c}

\hline
\multirow{2}{*}{Method} & \multicolumn{3}{c|}{W200} & \multicolumn{3}{c}{W100} \\ \cline{2-7}



& RMSE$\downarrow$ & MAE$\downarrow$ & $r\uparrow$ & RMSE$\downarrow$ & MAE$\downarrow$ & $r\uparrow$ \\ \hline

\multirow{2}{*}{Student} & 6.253 & 4.558 & 98.757 & 6.458 & 4.731 & 94.246 \\
& {\scriptsize$\pm$0.259} & {\scriptsize$\pm$0.162} & {\scriptsize$\pm$0.071} & {\scriptsize$\pm$0.229} &
{\scriptsize$\pm$0.158} & {\scriptsize$\pm$0.390} \\ \hline \hline

\multirow{3}{*}{Teacher1} & 5.977 & 4.402 & 98.800 & 6.375 & 4.700 & 94.676 \\
& {\scriptsize$\pm$0.209} & {\scriptsize$\pm$0.141} & {\scriptsize$\pm$0.061} & {\scriptsize$\pm$0.219} &
{\scriptsize$\pm$0.150} & {\scriptsize$\pm$0.332} \\
 & (0.001\textcolor{blue}{$\downarrow$}) & (0.026\textcolor{red}{$\uparrow$}) & (0.060\textcolor{red}{$\uparrow$}) & (0.054\textcolor{red}{$\uparrow$}) & (0.052\textcolor{red}{$\uparrow$}) & (0.124\textcolor{red}{$\uparrow$}) \\


\multirow{3}{*}{SCKD} & 5.788 & 4.240 & 98.866 & 6.183 & 4.559 & 95.136 \\
& {\scriptsize$\pm$0.170} & {\scriptsize$\pm$0.096} & {\scriptsize$\pm$0.051} & {\scriptsize$\pm$0.179} &
{\scriptsize$\pm$0.113} & {\scriptsize$\pm$0.335} \\
 & (0.153\textcolor{blue}{$\downarrow$}) & (0.111\textcolor{blue}{$\downarrow$}) & (0.078\textcolor{red}{$\uparrow$}) & (0.083\textcolor{red}{$\uparrow$}) & (0.037\textcolor{red}{$\uparrow$}) & (0.605\textcolor{red}{$\uparrow$}) \\

\hline

\multirow{3}{*}{Teacher2} & 5.488 & 4.048 & 98.969 & 5.447 & 4.092 & 96.468 \\
& {\scriptsize$\pm$0.227} & {\scriptsize$\pm$0.163} & {\scriptsize$\pm$0.090} & {\scriptsize$\pm$0.256} &
{\scriptsize$\pm$0.200} & {\scriptsize$\pm$0.294} \\
 & (0.106\textcolor{red}{$\uparrow$}) & (0.086\textcolor{red}{$\uparrow$}) & (0.058\textcolor{blue}{$\downarrow$}) & (0.073\textcolor{red}{$\downarrow$}) & (0.036\textcolor{red}{$\uparrow$}) & (0.162\textcolor{red}{$\uparrow$}) \\

\multirow{3}{*}{SCKD} & 5.842 & 4.337 & 98.879 & 6.302 & 4.622 & 95.609 \\
& {\scriptsize$\pm$0.163} & {\scriptsize$\pm$0.114} & {\scriptsize$\pm$0.046} & {\scriptsize$\pm$0.215} &
{\scriptsize$\pm$0.161} & {\scriptsize$\pm$0.294} \\ 
 & (0.058\textcolor{red}{$\uparrow$}) & (0.038\textcolor{red}{$\uparrow$}) & (0.022\textcolor{blue}{$\downarrow$}) & (0.013\textcolor{blue}{$\downarrow$}) & (0.002\textcolor{blue}{$\downarrow$}) & (0.011\textcolor{red}{$\uparrow$}) \\
 \hline

\multirow{3}{*}{Teacher3} & 5.787 & 4.231 & 98.989 & 5.788 & 4.235 & 94.810 \\
& {\scriptsize$\pm$0.282} & {\scriptsize$\pm$0.192} & {\scriptsize$\pm$0.080} & {\scriptsize$\pm$0.266} &
{\scriptsize$\pm$0.173} & {\scriptsize$\pm$0.392} \\
 & (0.164\textcolor{red}{$\uparrow$}) & (0.140\textcolor{red}{$\uparrow$}) & (0.140\textcolor{red}{$\uparrow$}) & (0.034\textcolor{red}{$\downarrow$}) & (0.047\textcolor{red}{$\uparrow$}) & (0.972\textcolor{blue}{$\downarrow$}) \\

\multirow{3}{*}{SCKD} & 6.129 & 4.512 & 98.872 & 6.344 & 4.642 & 95.268 \\
& {\scriptsize$\pm$0.230} & {\scriptsize$\pm$0.157} & {\scriptsize$\pm$0.055} & {\scriptsize$\pm$0.213} &
{\scriptsize$\pm$0.155} & {\scriptsize$\pm$0.357} \\ 
 & (0.012\textcolor{blue}{$\downarrow$}) & (0.019\textcolor{red}{$\uparrow$}) & (0.194\textcolor{red}{$\uparrow$}) & (0.078\textcolor{blue}{$\downarrow$}) & (0.048\textcolor{blue}{$\downarrow$}) & (0.184\textcolor{red}{$\uparrow$}) \\ \hline

\end{tabular}
}
\end{center}
\label{table:wae_ours2}
\end{table}

As explained in Table \ref{table:wae_ours2}, in most cases, AE teachers perform better than WAE teachers. We utilize these teachers in KD with the proposed method to analyze the effects of different teacher models trained using various encoder-decoder learning methods.
In the table, we compare the performance of SCKD using AE and WAE teachers. For both window sizes, WAE generally outperforms AE in terms of RMSE. Intuitively, when the window length is 100, in the cases of Teacher2 and Teacher3, SCKD with WAE teachers produces better student models than when using AE teachers. This supports the observation that a better teacher does not always lead to a superior student model \cite{cho2019efficacy, jeon2022kd}. In other words, even if a WAE teacher performs worse than an AE teacher, it can guide the training of a more effective student.
For Teacher1, the proposed method with WAE generates student models that outperform their respective teachers for both window lengths. This indicates that SCKD is effective when the architecture of the student is similar to that of the WAE teacher. This trend is also observed with AE teachers.
In the case of Teacher3, WAE yields better RMSE results than AE for both window lengths. Furthermore, in all cases, SCKD outperforms the baseline student model trained from scratch, demonstrating that SCKD is effective across teacher models trained with different learning approaches and architectures. Across all configurations, SCKD distills better student models with a window size of 200 compared to 100. This suggests that incorporating more temporal information (e.g., patterns and characteristics around inflection points) contributes to the training of more effective student models.

\section{Ablations and Sensitivity Analysis} \label{sec:ablations}

In this section, we analyze the effectiveness of the proposed method in various aspects, including correlation selection, layer selection, hyperparameters in distillation, and visualizations.

\subsection{Correlation Selection for Distillation} \label{temp_selc}
For the proposed method, we utilize correlations of features from a selected subset of channels, which is determined by hyperparameter $q$. Specifically, $q$ is the number of features used in knowledge transfer. Given a total of $c$ channels in the feature map, the number of selected indices, $q$, is computed by $[c/m]+1$, where $m$ denotes the regular interval of selected indices. These selected indices are denoted as $k$ in equation \eqref{eq_cmap}. In this section, we explore performance on different values of $q$.
To highlight the performance gaps between different approaches, we evaluate the proposed method on a smaller dataset that consists of approximately half the number of training and testing samples with window size of 200.

\begin{figure}[htb!]
\includegraphics[scale=0.35] {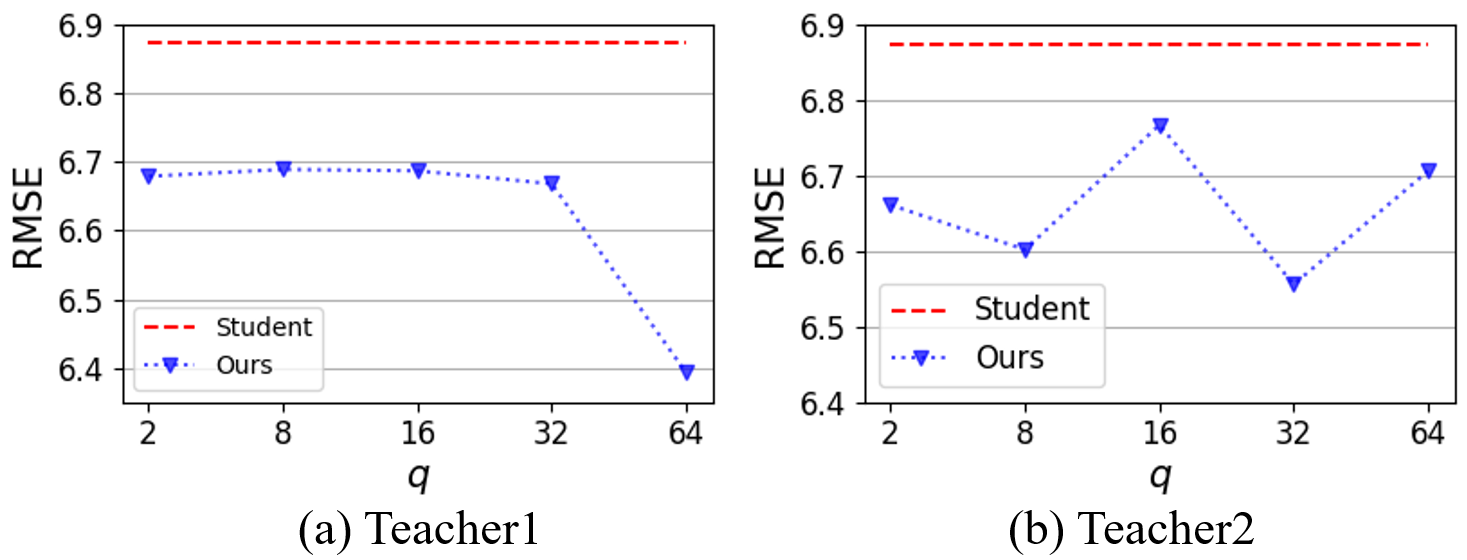} 
\centering
\caption{Analysis (RMSE ($\times10^{-2}$)) on the number of correlation maps in knowledge transfer.}
\label{figure:ch_num_select}
\end{figure}

We explore the effects of using different numbers of correlation maps, as shown in Fig. \ref{figure:ch_num_select}. To investigate the impact of correlation selection, we train the student using the middle representation without incorporating $\mathcal{L}_{KD_c}$.
For Teacher1, the result with $q$=64 shows better RMSE performance compared to $q$=32. However, the results from Teacher2 indicate that increasing the number of correlation maps does not always lead to better student performance. This suggests that excessive reliance on teacher features may hinder training of the student. Such dependency can restrict the student's ability to independently learn feature representations that align with its inherent architectural properties \cite{martin2021implicit, lan2019self}. This tendency becomes more prominent when the teacher and student architectures differ. In general, setting $q$=8 is recommended for the proposed method, as it is both effective and more efficient than using larger values of $q$.

\subsection{Layer Selection for Distillation} \label{layer_selec}
In KD, high- and mid-level layers have different impacts on training effectiveness \cite{jeong2022lightweight}. We explore the performance of the proposed method using different layer selections for encoder and decoder networks. As in the previous section, we use a smaller number of samples to better highlight the performance gaps between methods.
As shown in Table \ref{table:grf_inter1}, the combination of E2, Mid, and D1 generally yields the best results. For the proposed method, leveraging high-level features (i.e. $\{$E2, D1$\}$) outperforms the use of mid-level features (i.e. $\{$E1, D2$\}$). This suggests that incorporating too many intermediate layers in the distillation process may constrain the learning of a student, limiting its ability to independently acquire useful representations suited to its inherent architectural characteristics \cite{martin2021implicit, lan2019self}.


\begin{table}[htb!]
\centering
\caption{Performance (RMSE) for various combinations of layers in distillation. The check mark denotes selection of the features from the corresponding layer.}

\begin{center}
\begin{tabular}{c c c c c |c c c}

\hline
\multicolumn{2}{c}{Encoder} & & \multicolumn{2}{c|}{Decoder} & \multicolumn{3}{c}{Teacher Model} \\

E1 & E2 & Mid & D1 & D2 & Teacher1 & Teacher2 & Teacher3 \\ \hline

\multirow{2}{*}{--} & \multirow{2}{*}{--} & \multirow{2}{*}{\checkmark} & \multirow{2}{*}{--} & \multirow{2}{*}{--} & 6.518 & \textbf{6.470} & 6.641 \\
& & & & & {\scriptsize$\pm$0.276} &
{\scriptsize$\pm$0.244} & {\scriptsize$\pm$0.266} \\ \hline

\multirow{2}{*}{--} & \multirow{2}{*}{\checkmark} & \multirow{2}{*}{--} & \multirow{2}{*}{\checkmark} &\multirow{2}{*}{--} & 6.512 & 6.820 & 6.697 \\
& & & & & {\scriptsize$\pm$0.240} &
{\scriptsize$\pm$0.261} & {\scriptsize$\pm$0.277} \\ \hline

\multirow{2}{*}{--} & \multirow{2}{*}{\checkmark} &\multirow{2}{*}{\checkmark} & \multirow{2}{*}{\checkmark} & \multirow{2}{*}{--} & \textbf{6.380} & 6.523 & \textbf{6.562} \\
& & & & & {\scriptsize$\pm$0.236} &
{\scriptsize$\pm$0.244} & {\scriptsize$\pm$0.274} \\ \hline

 \multirow{2}{*}{\checkmark} & \multirow{2}{*}{--} &\multirow{2}{*}{\checkmark} & \multirow{2}{*}{--} & \multirow{2}{*}{\checkmark} & 6.477 & 6.478 & 6.754 \\
& & & & & {\scriptsize$\pm$0.270} &
{\scriptsize$\pm$0.270} & {\scriptsize$\pm$0.292} \\ \hline

 \multirow{2}{*}{\checkmark} & \multirow{2}{*}{\checkmark} &\multirow{2}{*}{\checkmark} & \multirow{2}{*}{\checkmark} & \multirow{2}{*}{\checkmark} & 6.414 & 6.603 & 6.594 \\
& & & & & {\scriptsize$\pm$0.239} &
{\scriptsize$\pm$0.262} & {\scriptsize$\pm$0.268} \\ \hline

\end{tabular}
\end{center}
\label{table:grf_inter1}
\end{table}

For further exploration, we compare the performance of the proposed method with AE and WAE across different window lengths, using only the middle representation in KD without additional intermediate layers from the encoder or decoder. As part of this extended investigation, we use the full dataset.
As explained in Table \ref{table:grf_inter2}, AE outperforms WAE when the window length is shorter. However, when the architecture of teacher differs significantly from that of the student, WAE performs better, particularly in the case of Teacher3 whose capacity is much larger than that of the student compared to other teachers. These results suggest that WAE is more effective when sufficient data is available and the teacher model is substantially different from the student.

\begin{table}[htb!]
\centering
\caption{RMSE results for different learning methods by leveraging middle representation in distillation.}

\begin{center}
\begin{tabular}{c | c |c c c}

\hline
 \multirow{2}{*}{Window} & \multirow{2}{*}{Method} & \multicolumn{3}{c}{Teacher Model} \\

& & Teacher1 & Teacher2 & Teacher3 \\ \hline

 \multirow{4}{*}{200} & \multirow{2}{*}{AE} & 5.940 & \textbf{5.808} & 6.160 \\
& & {\scriptsize$\pm$0.196} &
{\scriptsize$\pm$0.170} & {\scriptsize$\pm$0.259} \\ 

& \multirow{2}{*}{WAE} &  \textbf{5.728} & 5.847 & \textbf{6.029} \\
& & {\scriptsize$\pm$0.159} &
{\scriptsize$\pm$0.162} & {\scriptsize$\pm$0.212} \\ \hline

 \multirow{4}{*}{100} & \multirow{2}{*}{AE} & \textbf{6.100} & \textbf{6.261} & 6.398 \\
& & {\scriptsize$\pm$0.155} &
{\scriptsize$\pm$0.155} & {\scriptsize$\pm$0.204} \\ 

&  \multirow{2}{*}{WAE} & 6.145 & 6.346 & \textbf{6.375} \\
& & {\scriptsize$\pm$0.180} &
{\scriptsize$\pm$0.192} & {\scriptsize$\pm$0.213} \\ \hline

\end{tabular}
\end{center}
\label{table:grf_inter2}
\end{table}


\subsection{Hyperparameters for SCKD}
We analyze the proposed method with different hyperparameter choices for $\lambda_1$, $\lambda_2$, and $\lambda_3$, which are important factors for the objective of SCKD. We set $\lambda_1$, $\lambda_2$, and $\lambda_3$ to 1, 10, and 1, respectively, as a default setting. We measure performance of models in RMSE ($\times10^{-2}$). With the consistency of this section, we explore the proposed method with small dataset.

\begin{figure}[htb!]
\includegraphics[scale=0.35] {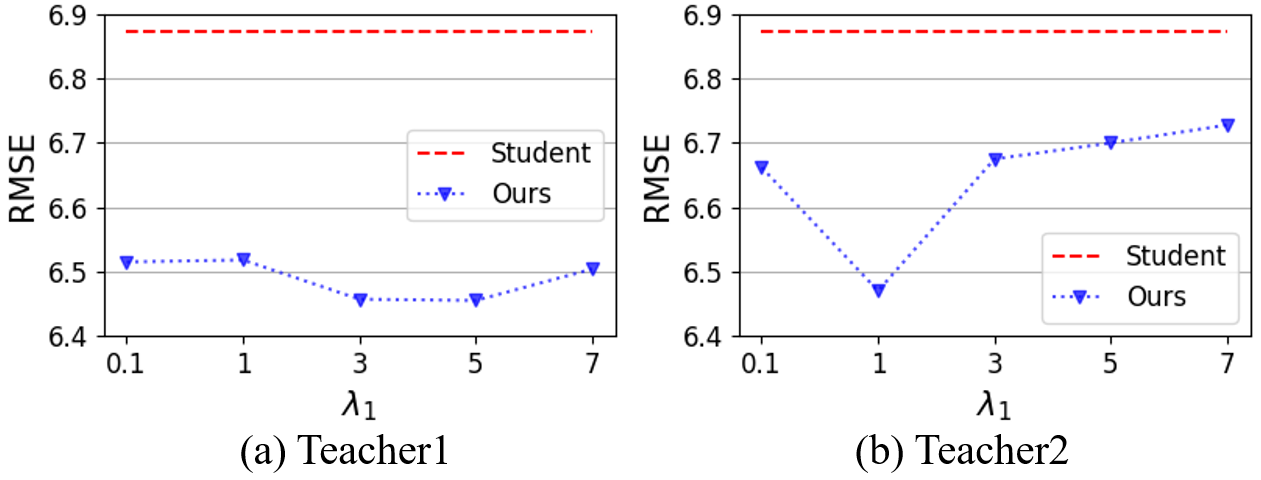}
\centering
\caption{Analysis on $\lambda_1$ of SCKD.}
\label{figure:lambda1}
\end{figure}

First, we analyze the proposed method with different values of the hyperparameter $\lambda_1$. As shown in Fig. \ref{figure:lambda1}, when the architecture of the teacher differs significantly from that of the student, setting $\lambda_1$ to 1 performs better than other values. Additionally, compared to Teacher1, the results for Teacher2 show larger performance gaps across different values of $\lambda_1$.

\begin{figure}[htb!]
\includegraphics[scale=0.35] {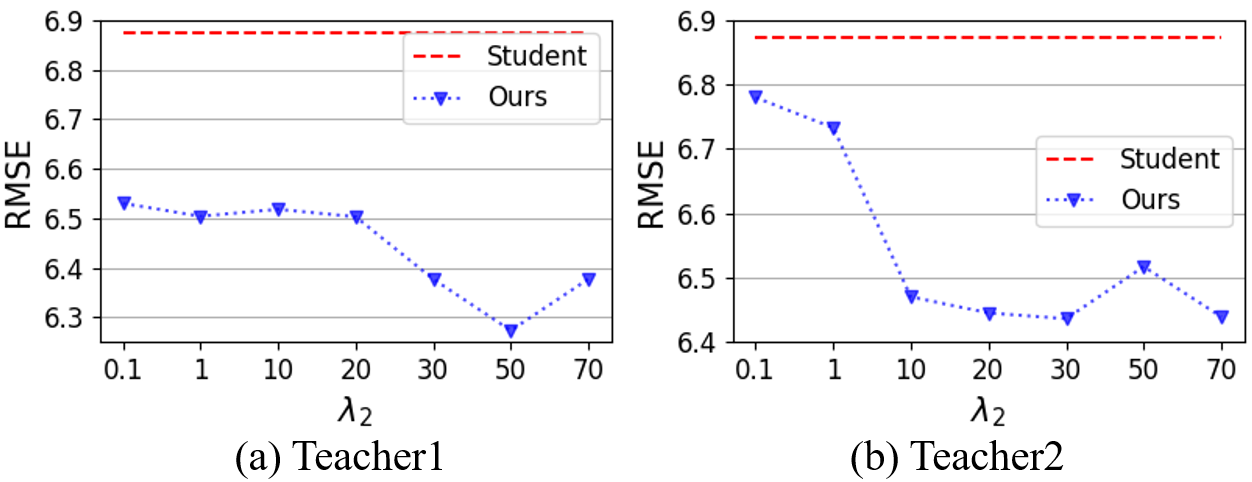}
\centering
\caption{Analysis on $\lambda_2$ of SCKD.}
\label{figure:lambda2}
\end{figure}

Second, we explore the performance for $\lambda_2$, a hyperparameter that influences the middle representation. As illustrated in Fig. \ref{figure:lambda2}, $\lambda_2$ values of greater than 20 outperform smaller values. However, values below 50 perform better than larger ones. When the full dataset is used for training, the proposed method performs better with a window size of 100 than with 200. This implies that middle representations can provide significant knowledge, especially when only a limited amount of information per sample is available during training.

\begin{figure}[htb!]
\includegraphics[scale=0.35] {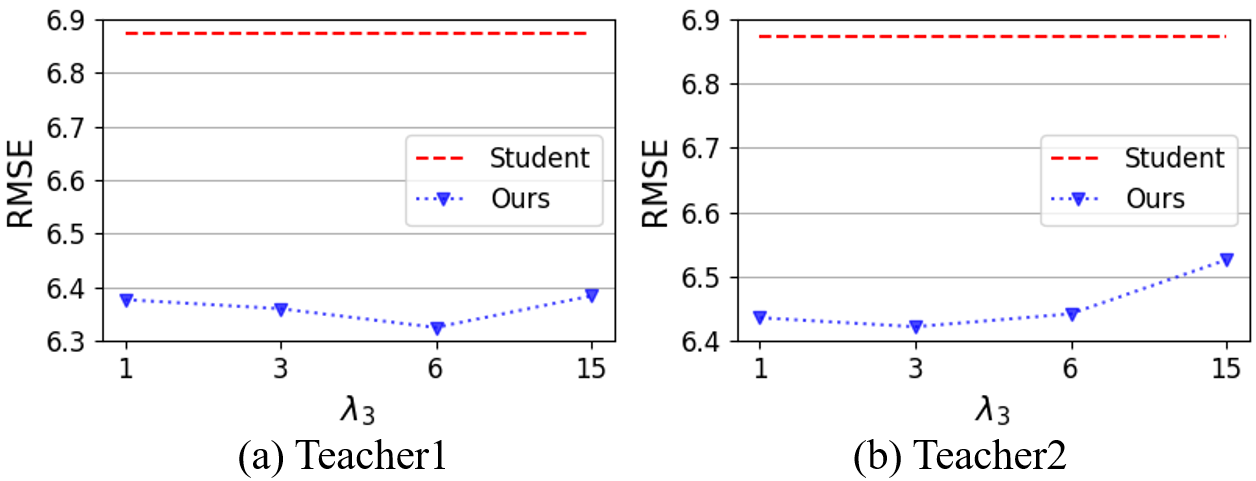}
\centering
\caption{Analysis on $\lambda_3$ of SCKD.}
\label{figure:lambda3}
\end{figure}

Finally, we investigate our method with different values of $\lambda_3$. We set $\lambda_2$ as 30, referring to results in Fig. \ref{figure:lambda2}. As described in Fig. \ref{figure:lambda3}, when $\lambda_3$ is set to 3 or 6, the proposed method shows better results than other settings. The $\lambda_3$ value is much less than $\lambda_2$, which can be approximated as $\lambda_2 \times0.1$. Based on these findings, we recommend setting $\lambda_3$ to approximately 0.1 of $\lambda_2$ for the proposed method.

\subsection{Visualization of Knowledge Distillation}
\subsubsection{Learned Features}

To investigate knowledge in distillation, we visualize feature relation maps for a more comprehensive analysis. We obtain $M$ of SP using equation \eqref{sp_eq}. Features from layers of an encoder ($\widehat{F}_{e} \in \mathbb{R}^{b\times c \times t \times w \times h}$) and a decoder ($\widehat{F}_{d} \in \mathbb{R}^{b\times c \times t}$) can be utilized to compute $M \in \mathbb{R}^{b\times b}$. We plot correlation maps using the previous method ($M$) and the proposed method ($G$) to compare the learned features across various models. Specifically, we utilize features from intermediate layers of an encoder (E2) and decoder (D1), which are near the middle representation (Middle).

\begin{figure*}[htb!]
\includegraphics[width=0.99\linewidth] {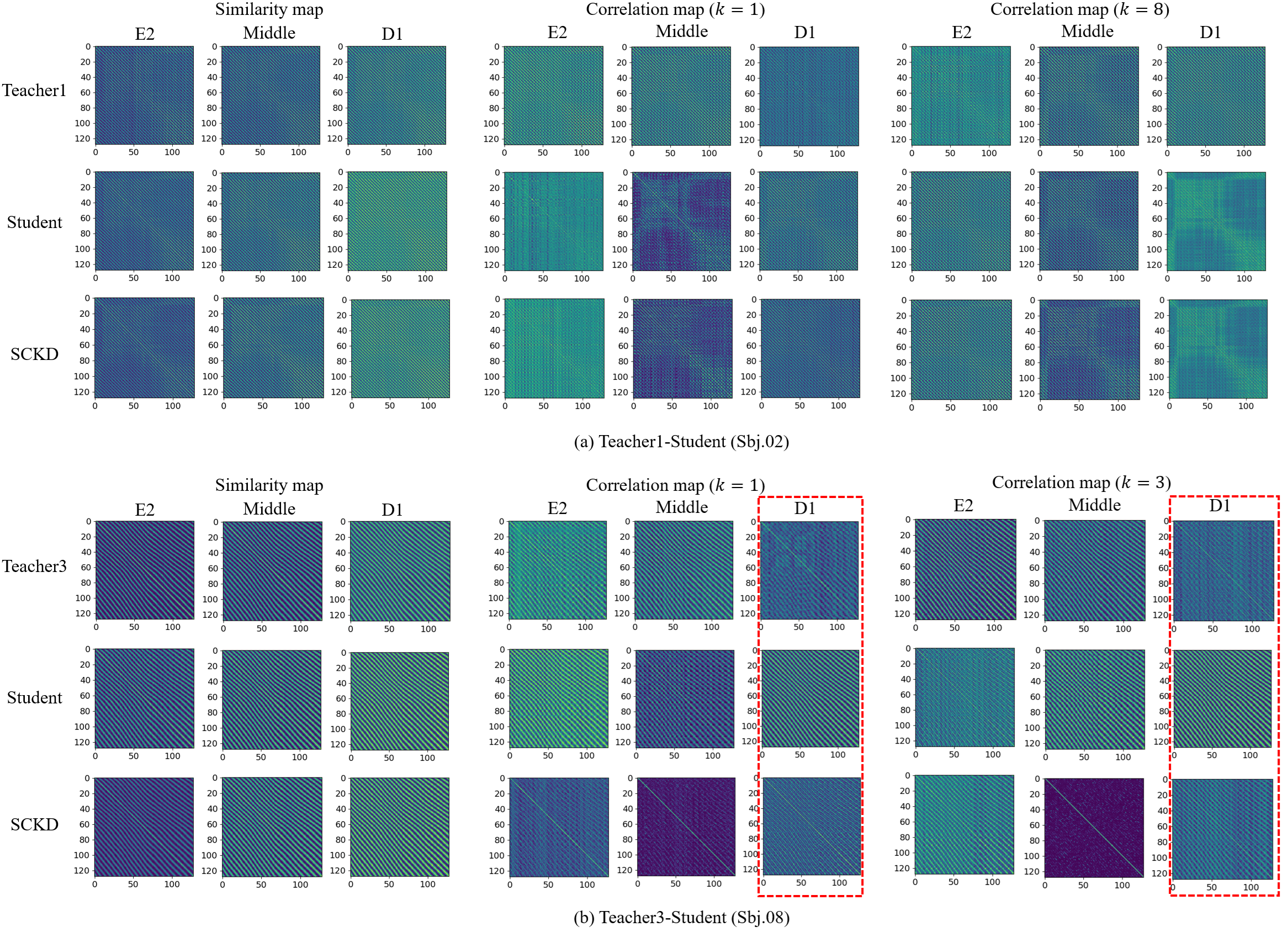}
\centering
\caption{Visualization of similarity and correlation maps ($M$ and $G$) for various models. $G$ is produced by our proposed method. $k$ denotes an index of channel for the feature to compute correlation map by the proposed method. Teacher and Student are models learned from scratch. SCKD denotes a student distilled by the proposed method.}
\label{figure:corr_map}
\end{figure*}

In Fig. \ref{figure:corr_map}, the similarity map, $M$, from SCKD is more similar to the map of the teacher compared to Student. This is particularly evident in the case of Teacher3, where the difference between the teacher and student models is more clearly shown than with Teacher1.

Since the process of estimation of time-series data is ultimately reliant on the decoder with 1D networks, it is important to infer relationships between features relevant to temporal information in the decoder. For Teacher1 and Teacher3, the D1 correlation maps from SCKD are more similar to those of the teacher than those from the student. For example, the student's results exhibit brighter maps, which deviate more from the teacher than the SCKD maps.

Remarkably, this tendency is more recognizable in the correlations maps, $G$, that are produced by the proposed method. The maps of D1 from Student are brighter than the distilled student from SCKD, which is less similar to the ones from the teacher. For Teacher3-Student case, this is more prominently visualized. Compared to the Student map from D1, the SCKD map is more similar to the teacher map across different $k$.
This is highlighted in the red dashed box in Fig. \ref{figure:corr_map}(b) when the teacher is more architecturally different from student.
Furthermore, these results show that each correlation map corresponding to different $k$ captures different relationships of features. With this, SCKD is effective even when teacher and student have different architectures and knowledge gap increases.
Therefore, 
Considering the previous experimental results, selecting more than one value of $q$, though not an excessive number, provides richer and complementary knowledge about feature relationships, thereby enhancing the effectiveness of the distillation process.

\subsubsection{Estimation Results}
\begin{figure}[htb!]
\includegraphics[scale=0.35] {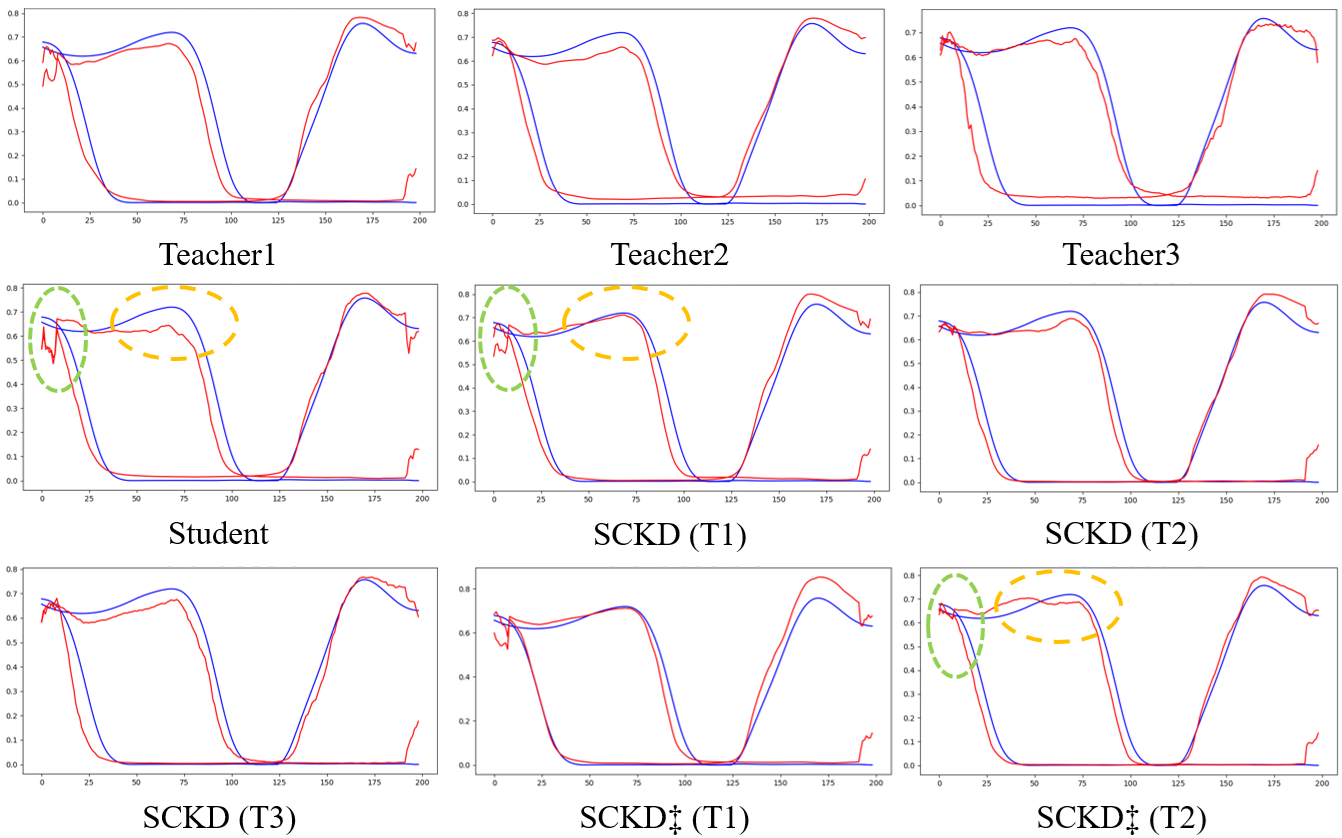}
\centering
\caption{Illustration of the estimation results from various models for the SW condition. Blue and red colored lines denote ground truth (GRF data) and estimation result, respectively. Brackets indicate a teacher model used in KD. $\ddagger$* denotes a student distilled by SCKD with WAE teacher. The yellow and green dashed lines clearly show the difference between the GRF estimated by Student and the proposed method, SCKD.}
\label{figure:est_result1}
\end{figure}

To determine the performance of various methods, we present the estimation results of different models in randomly selected samples in Fig. \ref{figure:est_result1}. In the figure, T1, T2, and T3 denote Teacher1, Teacher2, and Teacher3, respectively. SCKD indicates distilled student models by the proposed method with different teachers.
As shown in the first row, the teachers perform differently since the architectures of the models are different.
The results of students distilled by SCKD show fewer differences between blue and red lines, indicating less errors, compared to Student trained from scratch. This is more particularly at inflection points, which are highlighted in yellow and green dashed circles. Also, the results from SCKD$\ddagger$ (T2) shows more stable performance than others, which is presented with less gaps between GRF ground truth (red) and estimated result (blue) and is also shown with less variations at starting points and peaky points. Among the teacher models, T2 produces better results (lower RMSE and MAE scores) than T1 and T3, as explained in previous sections.

\begin{figure}[htb!]
\includegraphics[scale=0.375] {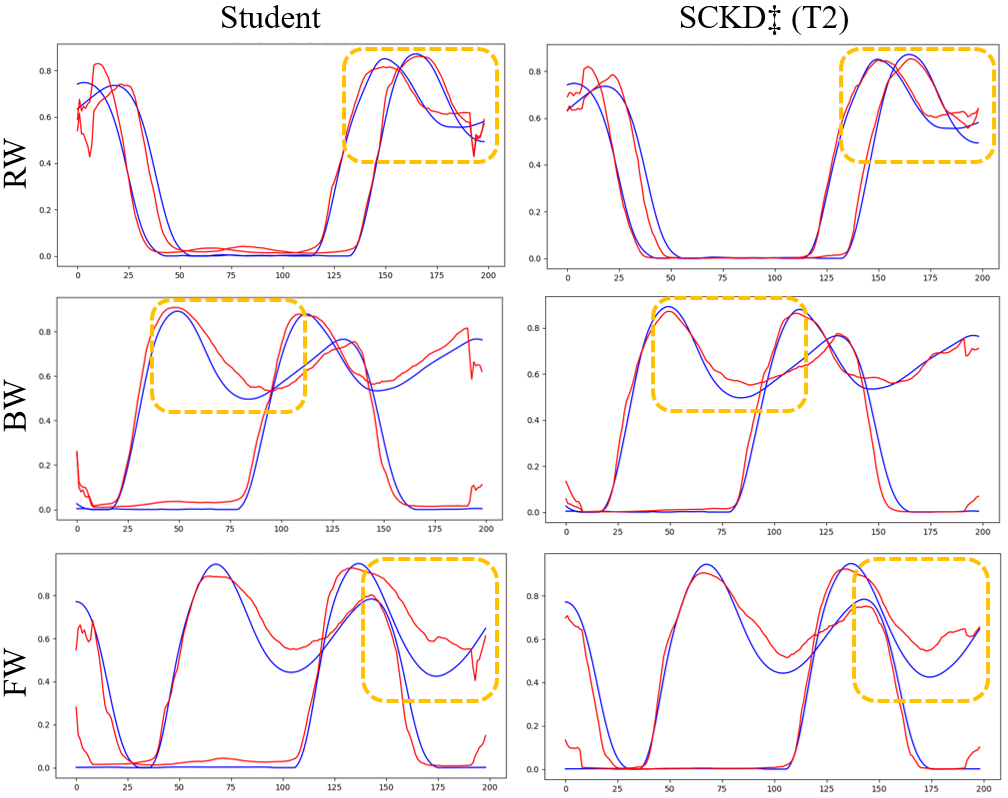}
\centering
\caption{Illustration of the estimation results from various models for different walking speed conditions. Blue and red colored lines denote ground truth (GRF data) and estimation result, respectively. $\ddagger$ denotes a student distilled by SCKD with WAE teacher. The yellow dashed lines clearly show the difference between the GRF estimated by Student and the proposed method, SCKD.}
\label{figure:est_result2}
\end{figure}

We also present results for Student learned from scratch and distilled student from SCKD on different walking conditions in Fig. \ref{figure:est_result2}. The student models were trained with the same settings of Section \ref{sec.4_2}, and evaluated on different levels of the velocities.
In all movement types, SCKD$\ddagger$ performs better than Student, with the improvement especially noticeable at high and low inflection points. This is highlighted by the yellow dashed boxes in the figure, which indicate that the difference between the estimation result (red) and the ground truth (blue) is smaller for SCKD than for Student.
Overall, it is noteworthy that SCKD estimates GRF with higher accuracy than Student trained from scratch for divserse activities.

\subsection{Model Reliability}
We compute the Expected Calibration Error (ECE) to measure model generalizability \cite{frank2015regression, naeini2015obtaining, guo2017calibration}, where ECE is commonly used as a metric to assess model reliability and reflects the degree of miscalibration in model confidence. We utilize various methods alongside this metric to evaluate the miscalibration of GRF estimates, comparing model predictions to ground truth measurements obtained from a treadmill. These evaluations are conducted over the entire dataset using a window length of 200, as described in Table \ref{table:ece}. 
These results aid in further explanation of the intrinsic nature of the subjects' biomechanics. For instance, student models from diverse methods tend to show higher reliability for right foot data compared to left foot data, implying that the subjects in this experiment exhibit asymmetries in their physical structures and movement behaviors.
On average results, the proposed method, SCKD with WAE T1 (denoted as SCKD$^\ddagger$), achieves the best performance. Among the AE-based teacher models, SCKD with T1 outperforms the baselines. Moreover, SCKD shows smaller reliability gaps between the two feet compared to other methods. This may be because SCKD consider both inter- and intra-temporal distributions in KD through the term $\mathcal{L}_{KD_c}$.
Among the teacher models, T1 performs best, likely due to its architectural similarity to the student model. This highlights that reliability scores and their rankings may differ from those based on RMSE. Based on these comprehensive observations, we conclude that SCKD effectively distills a student model that excels in both estimation accuracy and reliability.

\begin{table}[htb!]
\centering
\renewcommand{\tabcolsep}{1.3mm} 
\caption{ECE ($\%$) for various methods on GRF estimation with window size of 200. T1, T2, and T3 denote Teacher1, Teacher2, and Teacher3, respectively. $\ddagger$ denotes a teacher trained with WAE. \textbf{Bold} and \textcolor{red}{red} indicate best and second-best results, respectively.}

\begin{center}
\begin{tabular}{c |c c c || c| c c c}

\hline

Method & Left & Right & Avg. & Method & Left & Right & Avg.\\ \hline
Student & 2.011 & 1.654 & 1.832 & SCKD (T1) & \textbf{1.704} & \textcolor{red}{1.652} & \textcolor{red}{1.678} \\ \cline{1-4}

SP (T1) & 2.007 & 1.616 & 1.811 & SCKD (T2) & 2.188 & 1.971 & 2.080 \\
SP (T2) & 1.909 & 1.654 & 1.781 & SCKD (T3) & 2.000 & 2.011 & 2.004 \\ \hline 
DIST (T1) & 1.810 & 1.682 & 1.747 & SCKD (T1$^\ddagger$) &  \textcolor{red}{1.804} & \textbf{1.542} & \textbf{1.673} \\
DIST (T2) & 2.140 & 2.002 & 2.071 & SCKD (T2$^\ddagger$) & 2.127 & 2.080 & 2.104 \\ \cline{1-4}

\multicolumn{4}{c|}{} & SCKD (T3$^\ddagger$) & 2.064 & 1.914 & 1.989 \\ \cline{5-8}


\end{tabular}
\end{center}
\label{table:ece}
\end{table}








\subsection{Processing Time}\label{sec:computetime_chap}
We evaluate the processing time of different models using the all samples (approximately 7.6k data with 200 window) with a batch size of 1. The experiments were conducted on a desktop system equipped with an AMD Ryzen Threadripper PRO 5955WX 16-core and 32 threads with 4.58 GHz, 256GB memory (Device1), and a laptop system equipped with an Apple M2 chip with 8-core CPU and 8 GB memory (Device2). Since deep learning models typically require parallel threads and large-scale matrix operations, processing time was measured using a CPU, which is relatively more constrained in terms of parallelism. We used the teacher and student models from Table \ref{table:grf_w100w200}.
In Table \ref{tab:computation_time_merged}, the results show that Teacher3 takes the highest computational time, while student models take the lowest inference times. This clearly demonstrates the computational advantages of smaller models. Compared to KD baselines, the student model from SCKD achieves better accuracy in estimation.
Therefore, SCKD is effective in generating a compact model while preserving the comparable performance as a large model.

As shown in this table, the time difference is more prominently represented with Device2 as its specifications are significantly lower than those of Device1.
As device specifications decrease, bottlenecks arising from limited computational resources and constrained memory access become increasingly prominent, resulting in progressively larger variations in processing time. Consequently, performance inefficiencies between models are further amplified.

These results highlight the necessity of KD for model compression, particularly for deployment on resource-constrained devices.

\begin{table}[htb!]
\centering
\caption{Processing time of diverse models on CPU with different devices.}
\label{tab:computation_time_merged}

\begin{center}
{
\begin{tabular}{c|c|c|c|c|c|c|c}
\hline
\multicolumn{2}{c|}{\multirow{2}{*}{Model}} & \multicolumn{4}{c|}{Learning from scratch} & \multicolumn{2}{c}{KD} \\
\cline{3-8}
\multicolumn{2}{c|}{} & Teacher1 & Teacher2 & Teacher3 & Student & SemCKD & SCKD \\
\cline{3-6}
\hline
\multicolumn{2}{c|}{RMSE ($\times10^{-2}$)$\downarrow$} & 5.978& 5.382 & 5.623 & 6.253 & 6.197 & \textbf{5.784} \\
\hline
\multirow{2}{*}{Device1} & Total (sec) & 32.67 & 44.41 & 51.54 & \multicolumn{3}{c}{\textbf{18.34}} \\
 & Aver. (ms) & 4.18 & 5.69 & 6.60 &\multicolumn{3}{c}{\textbf{2.35}} \\
\hline
\multirow{2}{*}{Device2} & Total (sec) & 84.64 & 101.09 & 138.71 & \multicolumn{3}{c}{\textbf{38.48}} \\
 & Aver. (ms) & 10.83 & 12.94 & 17.76 &\multicolumn{3}{c}{\textbf{4.93}} \\
\hline
\end{tabular}
}
\end{center}
\end{table}

\section{Discussion} \label{sec:discussion}

As described in the previous sections, SCKD achieved robust results in generating a superior student model for GRF estimation and reliability. The estimation results from several models show perturbations near the starting and ending points of each sample. If a model uses larger size of padding in deep learning process, this issue would be alleviated. Training a model with inputs padded in early and ending indices may also help address this problem.
The performance of estimating heel strike and toe off positions can be improved by using customizable insole sensors. In detail, the insole sensor used in this paper was designed for a US men's shoe size 9.5 and included only partial sensor, which contributed to misalignment issues. Using sensors tailored to each individual could lead to better estimation results. Additionally, to improve temporal alignment, time warping methods can be applied \cite{jeon2023robust, lohit2019temporal}. Developing a temporal warping function for spatio-temporal data understanding in time-series estimation could be explored further as an extension of this study.

When processing movement related data, considering the generalizability of a model is crucial, especially when facing challenges in leveraging multimodal data (e.g., 3D and 1D CNNs) and accounting for personal variability. To address this, we evaluate the collected data using a leave-one-subject-out metric in various testing conditions, including different window lengths and number of data, learning strategies, model sensitivity, and visualized features and results.
In the data pre-processing stage, GRF data obtained from the force plates (measured in $N$) are normalized by each participant's weight, allowing us to express the results as a percentage of body weight. Specifically, a student model learned from scratch resulted in RMSE values of 7.07\% and 7.30\%, and MAE values of 5.15\% and 5.35\% of body weight for W200 and W100, respectively. For a student model distilled by SCKD using Teacher2 on W200, the RMSE and MAE in body weight are 6.54\% and 4.86\%, respectively.
With Teacher1 on W100 using SCKD, the RMSE and MAE in bodyweight are 6.90\% and 5.11\%, respectively. Therefore, the proposed SCKD method distills superior student models for GRF estimation compared to previous methods.


For future work, we plan to extend data collection to include a larger number of subjects. Furthermore, this framework can be extended to cover a broader range of activities and conditions, such as different surface types and inclines, foot position estimation with center of pressures, various environmental conditions (e.g., outdoor activities), and shoes made of different materials. These help further analysis on sensor aligned investigations and deeper understanding of personal conditions. We also would like to extend our work with time warping methods which can perform better with a larger range of walking speeds.

\section{Conclusion} \label{sec:conclusion}
In this paper, we proposed a novel framework, called Selective Correlation Based Knowledge Distillation (SCKD), which leverages a smaller number of selected features during distillation to extract effective relationships among feature elements considering temporal properties and improving efficiency. We developed a system to collect and construct datasets for GRF data from a treadmill and frame data from an insole sensor. The proposed method was evaluated using diverse metrics and various data configurations, incorporating different architectures of networks, learning approaches, and combinations of teacher-student models. Across all experiments, the proposed method showed robust performance, highlighting the importance of using selective features and accounting for output distribution during distillation. Our findings provide valuable guidance for developing more sophisticated distillation techniques tailored to wearable technology and gait analysis.

The insole sensor employed in this study features a fabric-like design, making it susceptible to warping during use. Additionally, its low data resolution provides limited information for analysis, resulting in degraded performance and restricting its application potentials. Variations in individual foot sizes further impact the accuracy of estimations. With more advanced sensors capable of capturing higher-resolution data, we anticipate improved estimation performance and more detailed data acquisition.

The proposed approach holds significant potential for integration with heterogeneous systems, facilitating the alignment of various types of information. This capability allows for the effective fusion of important features from multiple sensors, thereby enhancing the applicability of the framework across a wide range of environments and use cases. As part of future work, we plan to incorporate additional health-related sensors to develop models for the early detection and treatment of various diseases.
Additionally, we aim to advance methods that leverage both linear and non-linear feature metrics to better replicate the performance of a teacher model, particularly in capturing cross-modality characteristics.










\bibliographystyle{elsarticle-num} 
\bibliography{main}

\end{document}